%% file: main.tex
\documentclass{article}
\PassOptionsToPackage{numbers,sort&compress}{natbib}
\usepackage[preprint]{neurips_2025}

\usepackage[utf8]{inputenc}
\usepackage[T1]{fontenc}
\usepackage{amsmath,amssymb}
\usepackage{booktabs}
\usepackage{graphicx}
\usepackage{microtype}
\usepackage{needspace}
\usepackage{xcolor}
\usepackage{xspace}
\newcommand{\etal}{\emph{et al.}\xspace}
\definecolor{linkblue}{RGB}{34,77,116}
\usepackage[colorlinks,allcolors=linkblue]{hyperref}

\title{How Reward Gain Imbalance Shapes Ranking: Rethinking Rubric-Based Reward Aggregation}

\makeatletter
\hypersetup{pdftitle={\@title}}
\makeatother

\author{%
	Zhengyuan~Fan$^{1}$ \quad
	Zhonghua~Wu$^{1}$ \quad
	Tianyu~Cheng$^{1}$ \quad
	Qun~Chen$^{1}$\thanks{Corresponding author.} \\
	$^{1}$School of Computer Science, Northwestern Polytechnical University \\
	\texttt{\{fanzhengyuan,wuxhua,ctyu,chenbenben\}@nwpu.edu.cn}
}
\begin{document}
\maketitle
\input{sec/0_abstract}
\input{sec/1_intro}
\input{sec/2_relate}
\input{sec/3_method}
\input{sec/4_exp}
\input{sec/5_conclusion}


{\small
\bibliographystyle{ieeenat_fullname}
\bibliography{main}
}

\input{sec/X_suppl}
\end{document}

%% file: sec/0_abstract.tex
\begin{abstract}
	Rubric-based reinforcement learning constructs rewards for aligning large language models with human preferences, typically by equally aggregating continuous scores that a judge assigns to individual criteria. However, scoring heterogeneity across criteria can distort candidate rankings. Existing work has focused on improving judges, learning criterion weights, or refining aggregation, whereas the interaction between scoring heterogeneity and ranking remains insufficiently understood. We formalize this heterogeneity as \emph{reward gain imbalance}: each criterion enters the aggregated reward with its own reward gain, the gap between its expected scores on satisfied and unsatisfied responses. Counterintuitively, our experiments show that removing this imbalance, even with ground-truth labels, lowers ranking accuracy. To explain this, we derive an exact decomposition of the reward difference between any two candidates into a base satisfaction signal, a reward gain imbalance term, and a residual term, showing that reward gain imbalance can attenuate or amplify the base signal, and finding that misrankings usually concentrate on pairs where it attenuates. Guided by this analysis, we propose the adaptive \textbf{D}ual-\textbf{C}enter \textbf{C}lipping aggregation (DCC), which bounds each criterion's contribution to the reward difference instead of simply removing the imbalance. Using only the scores of the current candidate group, it determines a clipping interval that imposes a common upper bound on per-criterion score differences between candidates, limiting the excessive influence of extreme scores while preserving continuous judgment information. Because this interval is re-estimated online for every rollout group, the method is well suited to policy training with Group Relative Policy Optimization (GRPO). Finally, extensive experiments show that, among the compared methods, DCC achieves the best results on all ranking metrics across benchmarks and the best WritingBench score as a GRPO reward.
\end{abstract}

%% file: sec/1_intro.tex
\section{Introduction}
\label{sec:intro}
With the rapid advancement of large language models (LLMs), their use has become increasingly widespread, making it crucial to ensure that their outputs are both harmless and helpful. In recent years, substantial research efforts have been devoted to this goal. For example, reinforcement learning from human feedback (RLHF)\cite{RLHF1, RLHF2, RLHF3} aims to align model outputs with human preferences and prevent the generation of harmful responses. However, conventional reinforcement learning typically relies on relatively limited sources of reward signals, making it difficult to evaluate whether a response truly satisfies human needs across multiple dimensions. To address this limitation, recent studies\cite{RBRL1, RBRL2, RBRL3} have proposed incorporating rubrics that specify multiple evaluation criteria into reward signals, thereby imposing more precise constraints on model outputs. Over time, this research direction has evolved into a well-established framework, encompassing the construction of large datasets annotated with rubrics\cite{Openrubric, Healbench, rubrichub}, the training of reliable evaluation models\cite{pointrubric} based on rubrics, and the refinement of rubric design\cite{Adaptrule}.

Despite substantial progress in rubric-based reinforcement learning (RBR), including advances in judge models, evaluation criterion construction, criterion weighting, and reward aggregation, the problem of how to effectively transform individual criterion scores into an overall reward remains challenging. The crucial reason is scoring heterogeneity; that is to say, scores assigned to different criteria may vary in scale and distribution, as observed in previous studies\cite{fair1,RLRV2}, potentially leading to unreliable rewards when these scores are directly aggregated. Scoring heterogeneity stems from a judge's scoring performance at two levels: criteria differ in how ambiguous they are, so a judge would score some more decisively than others, and a judge's own ability to verify a criterion also varies across criteria. To address this issue, researchers have explored learning adaptive criterion weights\cite{check, weight1, weight2}, transforming score representations, and developing new aggregation strategies\cite{RBRV} to improve reward quality. However, existing studies have primarily focused on mitigating the effects of scoring heterogeneity through these approaches, while the underlying mechanisms by which such heterogeneity influences the relative ranking of candidate responses remain insufficiently understood. In particular, when both the judge and criterion weights are fixed, it remains unclear how numerical differences in continuous scores affect the final ranking and under what conditions these effects are beneficial or detrimental.

To this end, we investigate the sources of scoring heterogeneity across rubric criteria and how it affects reward rankings. Let $D_k$ denote the score assigned by the judge model to a candidate response for the $k$-th criterion, and let $Y_k \in \{0,1\}$ indicate whether the response truly satisfies it. Defining the reward gain of criterion $k$ as $\Delta_k=\mathbb{E}[D_k\mid Y_k=1]-\mathbb{E}[D_k\mid Y_k=0]$, each score can be written exactly as
\begin{equation}
	D_k = a_k + \Delta_k Y_k + \varepsilon_k,
	\label{eq:intro_score}
\end{equation}
where $a_k=\mathbb{E}[D_k\mid Y_k=0]$ and $\varepsilon_k$ denotes a residual term with zero conditional mean given $Y_k$. Consequently, under equal aggregation over the $K$ criteria with weight $1/K$, the satisfaction of criterion $k$ enters the final reward with coefficient $\Delta_k/K$ rather than $1/K$, a phenomenon we call \textbf{reward gain imbalance}. Equal aggregation weights therefore do not yield equal contributions. A natural remedy is to remove this imbalance: using ground-truth labels, we shift and rescale each criterion's scores so that unsatisfied responses average 0 and satisfied ones average 1 (by the mapping \(g\) as defined in Section~\ref{subsec:gainimbalance}). Unfortunately, our experiments show that this remedy may actually lower performance. As shown in Table~\ref{tab:ranking_accuracy}, removing the imbalance results in lower values of the Exact metric, the fraction of candidate groups ranked entirely correctly, on both datasets, partly because dividing by a small gain enlarges that criterion's residual.

\begin{table}[htbp]
	\centering
	\caption{The performance of Mistral-7B in terms of Exact (\%), \emph{before} \(\to\) \emph{after} giving all criteria the same reward gain with ground-truth labels.}
	\label{tab:ranking_accuracy}
	\small
	\begin{tabular}{lcc}
		\toprule
		Score Method & PointRubric & RubricHub-6 \\
		\midrule
		Raw score $D$ & 75.30 $\to$ 70.26 & 14.87 $\to$ 12.39 \\
		Binary score $\mathbf{1}[D \geq 0]$ & 31.41 $\to$ 30.70 & 2.30 $\to$ 3.01 \\
		$\operatorname{sigmoid}(D)$ & 74.58 $\to$ 71.70 & 13.45 $\to$ 12.57 \\
		\bottomrule
	\end{tabular}
\end{table}

To explain this observation, consider two candidates \(i\) and \(j\), where \(i\) satisfies more criteria than \(j\), and let \(S_i\) and \(S_j\) be their numbers of satisfied criteria. Substituting Equation~\ref{eq:intro_score} into their reward difference splits it exactly into three parts:
\begin{equation}
	R_i-R_j=B_{ij}+G_{ij}+E_{ij}.
	\label{eq:intro_decomposition}
\end{equation}
In Equation~\ref{eq:intro_decomposition}, the first part, \(B_{ij}=\bar\Delta(S_i-S_j)/K\), denotes the reward difference that would remain if every criterion had the same reward gain, namely the average gain \(\bar\Delta\). It depends only on how many more criteria \(i\) satisfies than \(j\), the quantity that defines the true ranking, so we call it the base signal; it favors \(i\) whenever the judge gives satisfied responses higher scores on average. The second part, \(G_{ij}\), comes from the unbalanced gains. It is not the gain of a criterion, but a difference between the two candidates:
\begin{equation}
	G_{ij}=\frac1K\sum_{k=1}^{K}\big[Y_{ik}(\Delta_k-\bar\Delta)-Y_{jk}(\Delta_k-\bar\Delta)\big],
	\label{eq:intro_g}
\end{equation}
where \(Y_{ik}=1\) if the candidate \(i\) satisfies the criterion \(k\) and 0 otherwise. Each candidate thus collects \(\Delta_k-\bar\Delta\), or how far its gain exceeds the average, over the criteria it satisfies. Since the criteria satisfied by both candidates would cancel, \(G_{ij}\) actually compares the reward gains of the criteria satisfied only by \(i\) with those satisfied only by \(j\). The third part, \(E_{ij}\), collects the residuals, that is, how far each score deviates from the judge's typical score for satisfied or unsatisfied responses.

Based on this reward decomposition, it becomes clear that reward gain imbalance does not necessarily hurt the ranking: \(G_{ij}>0\) widens the correct margin, whereas \(G_{ij}<0\) narrows it and may even reverse the order. Empirically, our experiments show that misrankings usually concentrate on the latter pairs: for instance, for Mistral-7B on PointRubric, the original scores correctly rank 99.00\% of pairs with $G_{ij}>0$ but only 86.44\% of those with $G_{ij}<0$. Guided by this theoretical analysis, we propose the adaptive \textbf{D}ual-\textbf{C}enter \textbf{C}lipping aggregation (DCC), which, rather than eliminating reward gain imbalance, bounds each criterion's contribution to \(G_{ij}\) and \(E_{ij}\) so that the base signal \(B_{ij}\) can determine more rankings. It thereby limits the excessive influence of extreme score differences while preserving useful information from continuous judgments. In summary, our main contributions are as follows:
\begin{itemize}
	\item \textbf{Reward gain imbalance.}
	We identify and analyze the implicit role of reward gain imbalance in rubric-based reward aggregation, pointing out that equal weights do not yield equal contributions, and show that the natural remedy of removing the imbalance, even with the help of ground-truth labels, cannot effectively improve ranking accuracy.
	
	\item \textbf{Reward Difference Decomposition.}
	We decompose the reward difference between any two candidates exactly into three components: the base satisfaction, reward gain imbalance and residual terms, which explains the inefficacy of eliminating the imbalance and locates frequent misrankings on pairs where reward gain imbalance attenuates the base signal.
	
	\item \textbf{Dual-Center Clipping Aggregation}.
	We propose DCC, a simple and efficient aggregation strategy. Among the compared methods, it achieves the best results on all ranking metrics across benchmarks and, as the reward in Group Relative Policy Optimization (GRPO)~\cite{shao2024deepseekmath}, the best WritingBench score.
\end{itemize}

%% file: sec/2_relate.tex
\section{Related work}
\label{sec:relatedwork}
\textbf{Rubric based Reinforcement Learning.} Reinforcement learning with verifiable rewards has greatly improved the reasoning of large language models in domains where correctness can be checked automatically~\cite{guo2025deepseekr1}. For open-ended tasks, rubric-based reinforcement learning replaces the verifier with natural-language criteria: an LLM judge evaluates a response against each criterion, and the judgments are combined into a scalar reward~\cite{RBRV, viswanathan2025checklists, huang2025rubricanchors}. Compared with holistic preference reward models~\cite{RLHF1}, such rewards are interpretable and less prone to over-optimization~\cite{zhang2026chasing}. Existing work mainly improves the rubric~\cite{zhang2026chasing, shen2026rrd} and the judge, which is typically a general LLM-as-a-judge~\cite{zheng2023judging} or a small model trained for rubric grading~\cite{pointrubric}; continuous scores are commonly derived from output-token probabilities~\cite{liu2023geval, zhang2025genverifier}, the type of score analyzed in this paper. Known judge biases, such as position and verbosity bias~\cite{zheng2023judging}, concern individual judgments. We instead take the rubric and the judge as given and show that, even when the judge separates satisfied from unsatisfied criteria correctly, the size of this separation can differ across criteria, so that equally weighted criteria contribute unequal gains.

\textbf{Score Aggregation as Reward.} Given the judge's scores on all criteria, an aggregation method turns them into a scalar reward. The default is linear aggregation, which sums or averages the raw scores~\cite{hayes2022practical, wu2023finegrained, viswanathan2025checklists}; when the task specifies criterion importance, it simply becomes a weighted sum. We regard weighting and aggregation as two separate levels: weights rescale each criterion's contribution before the summation, whereas the aggregation method determines how each score is mapped and combined. Weights may be assigned by rubric designers~\cite{RBRV, Healbench}, learned from preference data~\cite{RBRL1, wang2024armorm}, derived from correlations among criteria~\cite{shen2026rrd}, or adapted during training~\cite{moskovitz2024confronting, huang2026focal, fair1}, and any of them can be paired with any aggregation method. Beyond linear aggregation, nonlinear mappings are common. Binary aggregation thresholds each judgment and counts the satisfied criteria~\cite{RBRV, Healbench}; it sacrifices continuity and more strongly indicates a state of satisfaction or dissatisfaction, but this results in a large number of ties. Probability aggregation replaces each verdict with the judge's probability that the criterion is satisfied~\cite{RBRL1, zhang2025genverifier}; this fixed mapping saturates prematurely, pushing confident judgments to nearly 0 or 1, so probability aggregation degenerates toward binary aggregation. RLVR$^2$~\cite{RLRV2} departs further from summation: it keeps only within-group orderings and fuses criterion-wise Bradley--Terry utilities. Our Dual-Center Clipping aggregation (DCC) is another nonlinear mapping applied to each score before summation. Unlike binary and ordinal aggregation, it retains continuous judgment information; unlike probability aggregation, it saturates at two centers estimated from the current candidate group, so that only the amplitude beyond the group's typical low and high scores is removed. In GRPO~\cite{shao2024deepseekmath}, where each rollout forms such a group, the centers are re-estimated online, which makes the method well suited to this setting.

%% file: sec/3_method.tex
\section{Analysis of Reward Ranking}
\label{sec:method}
In this section, we first theoretically demonstrate the unequal contributions of individual scores to the aggregated reward, the phenomenon of reward gain imbalance. Then, we decompose the terms in reward signals, and analyze the role played by reward gain imbalance in reward ranking. Based on these analyses, Section~\ref{sec:dcc} then motivates and presents the proposed DCC approach.

\subsection{Derivation of Reward Gain Imbalance}
\label{subsec:gainimbalance}

We define a general scoring formulation, which is widely used for computing judge-model scores and can be expressed as follows:
\[D_{ik}=\log \pi_p(\mathrm{Yes}\mid x,y_i,r_{k})-\log \pi_p(\mathrm{No}\mid x,y_i,r_{k}),\]
where \(\pi_p\) denotes the judge model, \(x\) is the input prompt, \(y_i\) is the \(i\)-th candidate response, and \(r_k\) represents the \(k\)-th rubric. The resulting score \(D_{ik}\) is the log-probability ratio between affirmative and negative judgments, quantifying the judge model's confidence that \(y_i\) satisfies \(r_k\). If \(D_{ik}>0\), the judge model thinks that the current response satisfies the corresponding rubric; conversely, \(D_{ik}<0\) indicates the opposite. 

When aggregating, \(D_{ik}\) can be converted by a fixed function \(h\), such as the binary score \(h(d)=\mathbf{1}[d\geq 0]\) or the probability score \(h(d)=\operatorname{sigmoid}(d)\).
For convenience, we arrange the scores of the \(N\) candidate responses into a matrix \(D\in\mathbb{R}^{N\times K}\), whose entry \(D_{ik}\) is the score of candidate \(i\) on criterion \(r_k\). To describe the group as a whole, we draw a candidate uniformly at random from it and denote by \(D_k\) its score on criterion \(r_k\), so all expectations and probabilities below are averages over the \(N\) candidates of the current group. Therefore, the rubric-based reward of the drawn candidate is
\[R=\frac1K\sum_{k=1}^{K} D_{k},\]
where \(K\) is the number of rubrics. Extending the analysis to weighted aggregation only requires replacing \(1/K\) with specific criterion weights, without affecting our conclusions. Therefore, we use equal weights throughout the derivation for presentation simplicity.

Let \(Y_k\in\{0,1\}\) indicate whether the drawn candidate satisfies criterion \(r_k\), and let \(\theta_k=P(Y_k=1)\), i.e., the fraction of candidates that truly satisfy \(r_k\). Then, the expected score can be expressed as follows:
\begin{equation}
	\begin{aligned}
		\mathbb{E}[D_{k}]
		&= a_k(1-\theta_k)
		+(a_k+\Delta_k)\theta_k \\
		&= a_k+\Delta_k\theta_k,
	\end{aligned}
	\label{eq:expected_score}
\end{equation}
where $a_k=\mathbb{E}[D_{k}\mid Y_k=0]$ and $\Delta_k=\mathbb{E}[D_{k}\mid Y_k=1]-\mathbb{E}[D_{k}\mid Y_k=0]$. Based on the definition of \(R\) and Equation~\ref{eq:expected_score}, we can derive the expected reward over the current set of candidate responses as follows:
\begin{equation}
	\mathbb{E}[R] = \frac{1}{K}\sum_{k=1}^{K}\left(a_k + \Delta_k\theta_k\right).
	\label{eq:expected_reward}
\end{equation}

The term \(a_k\) contributes a constant to the expected reward, independent of the true satisfaction rate \(\theta_k\). Under equal weighting, the coefficient of \(\theta_k\) in \(\mathbb{E}[R]\) is \(\Delta_k/K\), indicating that different criteria can have unequal influences on the expected reward despite being assigned equal importance. More generally, when an importance weight \(w_k\) is assigned to criterion \(r_k\), its effective coefficient becomes \(w_k\Delta_k\). Consequently, the actual influence of each criterion on the expected reward depends not only on its assigned importance but also on its reward gain \(\Delta_k\). 

We have further conducted an empirical analysis to verify its existence and influence. The PointRubric\cite{pointrubric} dataset provides ground truth labels for each rubric, allowing us to readily estimate \(\Delta_k\) based on its definition. We then examine the mean and variance of these values over the train split, using the scores of Mistral-7B; the results are shown in Table~\ref{tab:gain_statistics}. The nonzero mean values in the table confirm the presence of imbalance reward gains \(\Delta_k\), while the nonzero variances indicate that its magnitude varies across different criteria.
\begin{table}[htbp]
	\centering
	\caption{Means and variances of \(\Delta_k\) under different scoring methods.}
	\label{tab:gain_statistics}
	\begin{tabular}{lcc}
		\toprule
		Scoring Method & Mean & Variance \\
		\midrule
		Raw score $D$
		& 21.9545 & 86.2175 \\
		Binary score $\mathbf{1}[D \geq 0]$
		& 0.7068 & 0.1363 \\
		Probability score $\operatorname{sigmoid}(D)$
		& 0.7071 & 0.1255 \\
		\bottomrule
	\end{tabular}
\end{table}

The definition of \(\Delta_k\) provides a theoretical quantification for reward gain imbalance. For each criterion, candidate responses can be divided into those that satisfy the criterion and those that do not, with \(\Delta_k\) measuring the difference in their expected scores. Although the judge model may assign higher scores to satisfied responses, the magnitude of this difference is not necessarily consistent across criteria. In particular, the model may exhibit different levels of confidence when making judgments under different criteria, resulting in varying values of \(\Delta_k\). Consequently, these differences lead to unequal influences on the expected reward, even when the criteria are assigned equal importance.

A straightforward remedy for reward gain imbalance is to eliminate such imbalance. To this end, we define a mapping function \(g\) that transforms each score into a calibrated score whose reward gain is the same for every criterion, as follows:
\[g_k(X_k)= \frac{X_k-a_k}{\Delta_k}, \]
where \(X_k=h(D_k)\) is the score under a chosen form \(h\), and \(a_k\) and \(\Delta_k\) are computed from \(X_k\) rather than from \(D_k\).
This transformation eliminates the effect of \(\Delta_k\) by giving every criterion equal gains.  We provide its proof in Section~\ref{sec:supp_gain} of the supplementary material.

Next, we apply \(g\) in the reward aggregation process and evaluate reward ranking accuracy under different transformations of \(D\). Notably, estimating both \(a_k\) and \(\Delta_k\) requires ground truth labels, making \(g\) difficult to estimate in settings without labeled data. We present the results of Mistral-7B on the PointRubric train split and RubricHub-6 in Table~\ref{tab:ranking_accuracy}. We present more experimental results in Section~\ref{sec:supp_calibration} of the supplementary material.

Counterintuitively, the results in Table~\ref{tab:ranking_accuracy} show that applying \(g(\cdot)\) to eliminate reward gain imbalance actually reduces the performance in terms of the Exact metric, even though \(g\) is estimated with ground-truth labels. This finding suggests that reward gain imbalance is not necessarily detrimental to reward ranking. This unexpected result motivates us to investigate the underlying relationship between reward gain imbalance and reward ranking, particularly how the two interact with each other. 

\subsection{Reward Signal Decomposition}
\label{subsec:rdd}

A response that satisfies more criteria should rank higher, where \(Y_{ik}=1\) means that response \(i\) satisfies criterion \(k\) as intended; for a negatively phrased criterion, such as one asking whether a response contains harmful content, this corresponds to a correct ``No''.

We now formalize the above ranking principle. Given a set of \(N\) candidate responses, let \(S_i=\sum_{k=1}^{K}Y_{ik}\) denote the number of criteria satisfied by a candidate \(y_i\). We define \(\mathcal P\) as the set of all required pairwise comparisons:
\[
\mathcal P
=\{(i,j)\mid S_i>S_j,\ i,j\in\{1,\ldots,N\}\}.
\]
For each pair \((i,j)\in\mathcal P\), the aggregated reward \(R_i\) should be strictly greater than \(R_j\). We therefore define the Exact indicator for a candidate group as:
\[
\operatorname{Exact}(\mathbf R,\mathbf S)
=\mathbf{1}\left[
\forall(i,j)\in\mathcal P,\ R_i>R_j
\right].
\]

Thus, \(\operatorname{Exact}=1\) if and only if every candidate pair is ranked correctly, so a single reversed pair breaks it. Our goal is to construct rewards from the scores \(D\) alone, without access to the labels \(Y_{ik}\), so that Exact holds for as many candidate groups as possible; it therefore suffices to examine the sign of \(R_i-R_j\) on each pair.

Given an arbitrary pair \((i,j)\in\mathcal P\), we examine how reward gain imbalance affects this sign. Using \(a_k\) and \(\Delta_k\) from Section~\ref{subsec:gainimbalance}, we define the residual \(\varepsilon_{ik}=D_{ik}-\mathbb{E}[D_{k}|Y_k=Y_{ik}]=D_{ik}-a_k-\Delta_kY_{ik}\), which gives the exact decomposition as follows: 
\begin{equation}
	D_{ik}=a_k+\Delta_kY_{ik}+\varepsilon_{ik},
	\label{eq:score_identity}
\end{equation}
with \(\mathbb{E}[\varepsilon_k\mid Y_k]=0\) by construction. Let \(\bar\Delta=\frac1K\sum_{k=1}^{K}\Delta_k\) and \(t_k=Y_{ik}-Y_{jk}\), so that \(\sum_k t_k=S_i-S_j>0\). Substituting Equation~\ref{eq:score_identity} into \(R_i-R_j\), the common baseline \(a_k\) cancels, and adding and subtracting \(\bar\Delta t_k\) yields
\begin{equation}
	\begin{aligned}
		R_i-R_j
		={}&\underbrace{\frac{\bar\Delta\,(S_i-S_j)}{K}}_{B_{ij}}
		+\underbrace{\frac1K\sum_{k=1}^{K}(\Delta_k-\bar\Delta)\,t_k}_{G_{ij}}\\
		&+\underbrace{\frac1K\sum_{k=1}^{K}(\varepsilon_{ik}-\varepsilon_{jk})}_{E_{ij}}.
	\end{aligned}
	\label{eq:decomposition}
\end{equation}

The above equation shows that, for any \(i\) and \(j\), the reward difference between the \(i\)-th and \(j\)-th candidate responses can be exactly decomposed into three components: \(B_{ij}\), \(G_{ij}\), and \(E_{ij}\), which we refer to as the base satisfaction signal, reward gain imbalance term, and the residual variation term, respectively.

For any given pair \((i,j)\in\mathcal P\), we have \(S_i>S_j\). We assume \(\bar{\Delta}>0\). This is a natural assumption for a reasonably discriminative judge that, on average across criteria, assigns higher expected scores to satisfied responses than to unsatisfied ones. Under this assumption, the base satisfaction signal \(B_{ij}\) is positive and favors the desired ordering. However, the reward gain imbalance term, \(G_{ij}\), and the residual variation term, \(E_{ij}\), may each be positive or negative. By the exact decomposition shown in Equation~\ref{eq:decomposition}, the pair is correctly ranked if and only if \(B_{ij}+G_{ij}+E_{ij}>0\). The reward gain imbalance term may contribute positively or negatively to the pairwise reward margin, while the final ordering depends on the combined effect of all three terms.

Our empirical analysis has also confirmed this analysis: using Mistral-7B on the PointRubric train split, the original scores correctly rank 99.00\% of required pairs with \(G_{ij}>0\) but only 86.44\% of those with \(G_{ij}<0\). The experimental results show that unsurprisingly, misrankings often concentrate where reward gain imbalance attenuates the base signal. See Section~\ref{sec:supp_gsign} of the supplementary material for more detailed experimental results.

\section{\texorpdfstring{Dual-Center Clipping Aggregation}{Dual-Center Clipping Aggregation}}
\label{sec:dcc}

In this section, we first motivate the design of the DCC strategy, and then formally present the technical details of DCC. 

\subsection{Motivation}
\label{subsec:implications}

The decomposition explains why the equalizing normalization \(g\) lowers the ranking accuracy in Table~\ref{tab:ranking_accuracy}. After \(g\), all gains equal 1, so \(G_{ij}^{(g)}=0\). However, \(g\) also divides the residuals by \(\Delta_k\), so the smaller a criterion's gain, the more its residual is scaled up relative to those of the other criteria. Eliminating reward gain imbalance therefore shifts the problem from \(G_{ij}\) to \(E_{ij}\).

Since pairs with \(G_{ij}>0\) are ranked correctly more often than those with \(G_{ij}<0\), one might instead try to enlarge \(G_{ij}\). This does not work either: the sign of \(G_{ij}\) depends on which candidate satisfies the high-gain criteria, so a large gain helps a pair when the better candidate satisfies that criterion but hurts another pair when the worse candidate does, and a gain that is too large lets a single criterion override \(B_{ij}\). The desired aggregation therefore keeps \(G_{ij}\) moderate: it keeps the positive base signal \(B_{ij}\) while bounding how much any single criterion can contribute to \(G_{ij}+E_{ij}\).

\begin{figure}[htbp]
	\centering
	\includegraphics[width=\linewidth]{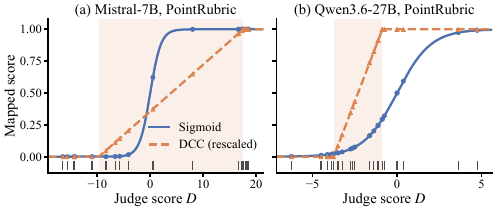}
	\caption{Visualization of score mappings of sigmoid and DCC on two real candidate groups: dots denote scores, and the band denotes DCC's clipping interval; DCC is rescaled to \([0,1]\) within the group, which leaves rankings unchanged. In (a) for Mistral-7B, the sigmoid maps 23 of 24 scores to nearly 0 or 1; in (b) for Qwen3.6-27B, most scores are negative; the sigmoid maps 18 of 24 below 0.3, while DCC sets its interval from this group's own scores, so the scores stay spread out.}
	\label{fig:sigmoid_defect}
\end{figure}

Binary aggregation achieves this goal in the ideal case: when every binary judgment matches \(Y_{ik}\), all gains equal 1, both \(G_{ij}\) and \(E_{ij}\) vanish, and \(R_i-R_j=(S_i-S_j)/K\). Unfortunately, real judges make errors, and binary aggregation then gives many candidates the same score, which weakens the continuity of the reward signal. Probability aggregation, \(\operatorname{sigmoid}(D)\), removes this discontinuity, but most scores fall in its saturated regions near 0 or 1, where they differ so little that their sum rarely changes the ordering (Figure~\ref{fig:sigmoid_defect}a). Its midpoint is also fixed at zero, while judge scores are usually not symmetric around zero: in a group with mostly negative scores, nearly all scores map close to 0 and the reward differences become weak, even though some of them are among the highest in the group (Figure~\ref{fig:sigmoid_defect}b; more results in Section~\ref{sec:supp_sigmoid} of the supplementary material).

These observations suggest two properties for score transformation: (1) a score should be interpreted relative to the other scores in the current group rather than on an absolute scale; and (2) it should be bounded on both sides, since without an upper bound a lower-ranked candidate can overtake through one excessively high score, and without a lower bound a higher-ranked candidate can be pushed down by one excessively low score. These two properties motivate DCC.

\subsection{DCC Aggregation Strategy}
\label{subsec:dcc_method}

Following the two properties in Section~\ref{subsec:implications}, the adaptive DCC strategy keeps the bounded structure of binary aggregation but replaces its two fixed values with two centers estimated from the current candidate group, so that every score is bounded while scores inside the interval keep their differences. We first describe DCC and then analyze how it serves this goal.

\vspace{0.05in}
\noindent\textbf{Center estimation:} we estimate the centers of the low- and high-score clusters in the current candidate group, denoted by \(c_-\) and \(c_+\). They serve as clipping endpoints shared across criteria, so the interval adapts to the group's score distribution.

Specifically, for the current prompt, we sort the \(NK\) finite entries of the score matrix \(D=(d_{ik})\) into \(d_{(1)}\leq\cdots\leq d_{(NK)}\), where parentheses indicate sorted positions. We enumerate splits \(s\in\{1,\ldots,NK-1\}\) for which \(d_{(s+1)}>d_{(s)}\), partitioning the sorted scores into a lower and an upper cluster. For each split, we use the median of each cluster as its center:
\begin{equation}
	\begin{aligned}
		c_-^{(s)}&=\operatorname{median}(d_{(1)},\ldots,d_{(s)}),\\
		c_+^{(s)}&=\operatorname{median}(d_{(s+1)},\ldots,d_{(NK)}).
	\end{aligned}
	\label{eq:dcc_split_centers}
\end{equation}
We use medians because they minimize the sum of absolute deviations within each cluster and are less sensitive to extreme score magnitudes compared with means. Accordingly, we evaluate each split by its total absolute deviation:
\begin{equation}
	\begin{aligned}
		\operatorname{cost}(s)={}&\sum_{m=1}^{s}|d_{(m)}-c_-^{(s)}|\\
		&+\sum_{m=s+1}^{NK}|d_{(m)}-c_+^{(s)}|.
	\end{aligned}
	\label{eq:dcc_split_cost}
\end{equation}
We choose the valid split \(s^\star\) with the smallest \(\operatorname{cost}(s)\), retaining the smallest \(s\) when costs tie, and set \(c_-=c_-^{(s^\star)}\) and \(c_+=c_+^{(s^\star)}\). If no valid split exists, all scores are equal, and we set \(c_-=c_+=d_{(1)}\). Note that such a scan implements exact one-dimensional two-medians fitting without ground-truth labels.

The total time complexity of DCC is \(O(NK\log(NK))\) per prompt, making the procedure suitable for online reward aggregation in GRPO\cite{shao2024deepseekmath}.  During training, the centers are therefore re-estimated on every rollout group, so the interval keeps pace with the judge's scores as the policy changes; Section~\ref{subsec:grpo} examines the effect on policy training.

\vspace{0.05in}
\noindent\textbf{Reward aggregation:} we form the clipped score matrix \(h\) by applying the same interval \([c_-,c_+]\) to every entry of \(D\):
\begin{equation}
	h_{ik}=\min\{c_+,\max\{c_-,d_{ik}\}\}.
	\label{eq:dcc_clipping}
\end{equation}
The scores inside the interval remain unchanged, while the scores outside it are replaced by the nearest endpoint. The shared interval bounds every per-criterion score difference between candidates by \(c_+-c_-\).

Given an optional set of criterion weights, \(w_k\), with a nonzero sum, the final reward is
\begin{equation}
	R_i^{\mathrm{DCC}}=\frac{\sum_{k=1}^{K}w_kh_{ik}}{\sum_{k=1}^{K}w_k}.
	\label{eq:dcc_reward}
\end{equation}
In case of equal weights, we set \(w_k=1\), recovering \(R_i^{\mathrm{DCC}}=\frac1K\sum_k h_{ik}\). Note that weights only enter the final averaging step and do not affect the centers. The entire procedure is applied independently to each prompt in a batch.

\vspace{0.05in}
\noindent\textbf{Analysis:} we now connect clipping to the reward signal decomposition as shown in Equation~\ref{eq:decomposition}. Fix a candidate group and its fitted centers, and consider equal criterion weights. We use the superscript \((h)\) for quantities recomputed from the clipped scores in Equation~\ref{eq:dcc_clipping}. In the ideal case, suppose \(c_-<c_+\) and, for every criterion, \(\mathbb{E}[h_k\mid Y_k=0]=c_-\) and \(\mathbb{E}[h_k\mid Y_k=1]=c_+\). Since \(h_k\in[c_-,c_+]\), these endpoint expectations imply \(h_{ik}=c_-+(c_+-c_-)Y_{ik}\) for every candidate \(i\). Therefore, all gains equal \(c_+-c_-\), and all residuals vanish. Consequently, \(G_{ij}^{(h)}=E_{ij}^{(h)}=0\), and every required pair satisfies
\begin{equation}
	\begin{aligned}
		R_i^{\mathrm{DCC}}-R_j^{\mathrm{DCC}}
		&=B_{ij}^{(h)}\\
		&=\frac{(c_+-c_-)(S_i-S_j)}{K}>0.
	\end{aligned}
	\label{eq:dcc_ideal_margin}
\end{equation}

However, in practice, the fitted centers do not necessarily coincide with these conditional means. Nevertheless, clipping is non-expansive and satisfies
\begin{equation}
	|h_{ik}-h_{jk}|\leq\min\{|d_{ik}-d_{jk}|,c_+-c_-\},
	\label{eq:dcc_contraction}
\end{equation}
which caps each criterion's contribution to the reward difference. This bound alone does not guarantee better ranking, since it also compresses the score differences that carry the base signal; what it ensures is that no single score, however extreme, changes the reward difference by more than \((c_+-c_-)/K\). If two candidates share the same satisfaction state on a criterion and their scores are clipped to the same endpoint, their residual difference on that criterion becomes zero.

It is worth pointing out that when the two candidates satisfy different criteria, clipping may change how their respective scores add up. For a pair with \(G_{ij}<0\), a high-gain criterion satisfied only by \(j\) may give it a large enough advantage to outweigh several smaller gaps favoring \(i\). If \(j\)’s large advantage comes from extreme scores, while \(i\)’s smaller advantages come from scores inside the clipping interval, DCC clips the former and leaves the latter unchanged. \(i\)’s advantages can then add up to put it ahead, even if $G_{ij}^{(h)}$ remains negative. On PointRubric and RubricHub-6 with both judges, 58 of the 74 pairs corrected by DCC had \(G_{ij}<0\) before clipping (Table~\ref{tab:supp_gsign} in supplementary material).

%% file: sec/4_exp.tex
\section{Experiments}
\label{sec:exp}

Our empirical study consists of two parts. The first and primary part (the Subsection~\ref{subsubsec:results}) studies how different aggregation methods affect reward ranking. The second part (the Subsection~\ref{subsec:grpo}) serves as auxiliary validation: we apply these aggregation methods to GRPO and examine their effect on the fine-tuned models. The ablation study is given in Section~\ref{sec:supp_ablation} of the supplementary material.
\begin{table}[htbp]
	\centering
	\caption{Ranking accuracy (\%) comparison of different aggregation methods on the PointRubric train split and RubricHub-6, with equal criterion weights.}
	\label{tab:main_ranking}
	\small
	\begin{tabular}{@{}lrrrrrr@{}}
		\toprule
		& \multicolumn{3}{c}{PointRubric (train)} & \multicolumn{3}{c}{RubricHub-6} \\
		\cmidrule(lr){2-4} \cmidrule(l){5-7}
		Method & Exact & Pairwise & Top-1 & Exact & Pairwise & Top-1 \\
		\midrule
		\multicolumn{7}{@{}l}{\emph{Mistral-7B}} \\
		Original $D$ & 75.30 & 93.49 & 87.05 & 14.87 & 86.28 & 72.21 \\
		Binary & 31.41 & 74.11 & 68.07 & 2.30 & 73.81 & 54.14 \\
		Probability & 74.58 & 93.45 & 89.93 & 13.45 & 85.97 & 71.68 \\
		RLVR$^2$ & 65.95 & 90.79 & 85.17 & 13.81 & 85.06 & 68.50 \\
		DCC (ours) & \textbf{77.22} & \textbf{94.36} & \textbf{90.17} & \textbf{16.11} & \textbf{86.34} & \textbf{73.63} \\
		\midrule
		\multicolumn{7}{@{}l}{\emph{Qwen3.6-27B}} \\
		Original $D$ & 84.89 & 96.88 & 96.16 & 36.99 & 93.41 & 90.97 \\
		Binary & 62.83 & 91.02 & 95.04 & 17.17 & 88.69 & 87.46 \\
		Probability & 85.13 & 96.88 & 96.64 & 36.11 & 93.22 & 90.62 \\
		RLVR$^2$ & 84.65 & 96.79 & 95.92 & 38.76 & 93.53 & 91.33 \\
		DCC (ours) & \textbf{85.37} & \textbf{97.02} & \textbf{97.36} & \textbf{42.12} & \textbf{93.96} & \textbf{92.74} \\
		\bottomrule
	\end{tabular}
\end{table}

\subsection{Experimental Setup}
\label{subsec:setup}

\noindent\textbf{Datasets for reward ranking:} we use the train and test splits of the public PointRubric dataset\cite{pointrubric}. Both splits provide ground-truth labels for each criterion, so both can be used to compute ranking accuracy. In an additional experiment on learnable criterion weights, we fit the weights on the train split and evaluate them on the test split. We also construct RubricHub-6 from \href{https://huggingface.co/datasets/sojuL/RubricHub_v1/blob/main/sft_RuFT/rurbichub_v1_6samples_156k_sft_data.parquet}{RubricHub}\cite{rubrichub}: we take its prompts with 16 criteria, regenerate candidate responses with GPT-6 Astra, and keep six responses per prompt, chosen to cover different satisfaction levels and ordered by the number of satisfied criteria. Its labels come from model-based review rather than human annotation.

\noindent\textbf{Judge models:} we use \href{https://huggingface.co/mistralai/Mistral-7B-Instruct-v0.3}{Mistral-7B-Instruct-v0.3}\cite{mistral} and \href{https://huggingface.co/Qwen/Qwen3.6-27B}{Qwen3.6-27B}\cite{qwen3.6-27b} as judges. Both produce the scores, \(D_{ik}\), as defined in Section~\ref{subsec:gainimbalance} under the same prompt template, and all the aggregation methods operate on the same scores.

\noindent\textbf{Compared methods:} all methods average the same scores \(D_{ik}\) over criteria with equal weights and differ only in how each score is transformed. Original \(D\) uses the raw scores, Binary uses \(\mathbf{1}[D_{ik}\geq 0]\), and Probability uses \(\operatorname{sigmoid}(D_{ik})\). RLVR\(^2\)~\cite{RLRV2} converts scores into pairwise win/tie/loss outcomes, fits criterion-wise Bradley--Terry utilities, and averages them to obtain the aggregated score.

\noindent\textbf{Metrics:} a candidate's ground-truth rank is given by its number of satisfied criteria \(S_i\), and pairs with equal \(S_i\) impose no constraint. We report three metrics, all averaged as percentages. The \emph{Exact} is the fraction of candidate groups whose rewards reproduce the full ground-truth order. The \emph{Pairwise} is the fraction of all strictly ordered pairs, pooled across groups, that are ranked correctly. The \emph{Top-1} measures whether the highest-reward candidate is a truly best one; when several candidates tie for the highest reward, the group receives the fraction of them that are truly best.

\subsection{Main results}
\label{subsubsec:results}

We run both judges on each dataset with the same vLLM\cite{vllm} inference configuration and the same scoring prompt; the implementation details are given in Section~\ref{sec:supp_inference} of the supplementary material. Table~\ref{tab:main_ranking} reports the main comparative results. It can be observed that DCC achieves the highest Exact in all four settings, exceeding the best competing method by 1.92 and 1.24 percentage points on PointRubric and RubricHub-6 respectively under Mistral-7B, and by 0.24 and 3.36 percentage points respectively under Qwen3.6-27B. The judge also matters: on the same candidates, replacing Mistral-7B with Qwen3.6-27B raises the Exact of the original scores D from 75.30\% to 84.89\% on PointRubric and from 14.87\% to 36.99\% on RubricHub-6, since a more reliable judge scores the criteria more accurately and the three terms in Equation~\ref{eq:decomposition} thus more often add up in agreement with the true ordering. Relative to the original scores, most pairs corrected by DCC have \(G_{ij}<0\), where reward gain imbalance attenuates the base signal. More detailed experimental results are presented in Section~\ref{sec:supp_gsign} of the supplementary material.

Since a weight \(w_k\) enters the reward as \(w_k\Delta_k\) (Section~\ref{subsec:gainimbalance}), its effect may differ from the importance it encodes; to test this, we compare equal, official, and learned weights on the PointRubric test split. The learned weights come from a small network that maps the score matrix \(D\) and the criterion types of a prompt to one positive weight per criterion; following Mu \etal\cite{RBRL1}, it is trained on the train split with a pairwise hinge loss (Section~\ref{sec:supp_weights} of the supplementary material).

The evaluation results presented in Table~\ref{tab:criterion_weights} show that using criterion importance directly as weights is not reliable: compared with equal weights, official weights lower Exact in four of the ten settings and leave it unchanged in two, consistent with our analysis. In comparison, learned-weight aggregation performs more stably, raising Exact in nine of the ten settings, and the only decrease is 0.19 percentage points. However, this performance gain has a cost, since learning weights requires a labeled dataset and network optimization, and they are prone to overfitting. We provide more evaluation results in Section~\ref{sec:supp_weights} of the supplementary material.

\begin{table}[htbp]
	\centering
	\caption{Performance comparison in terms of Exact (\%) on the PointRubric test split under different criterion weights: the learned weights use the judge scores and criterion types as input, and the official weights are provided by the benchmark.}
	\label{tab:criterion_weights}
	\small
	\setlength{\tabcolsep}{2.5pt}
	\begin{tabular}{lcccccc}
		\toprule
		& \multicolumn{3}{c}{Mistral-7B} & \multicolumn{3}{c}{Qwen3.6-27B} \\
		\cmidrule(lr){2-4} \cmidrule(lr){5-7}
		Method & Equal & Official & Learned & Equal & Official & Learned \\
		\midrule
		Original $D$ & 75.82 & 75.82 & 76.20 & \textbf{81.57} & 80.81 & 81.38 \\
		Binary & 30.71 & 32.05 & 32.82 & 56.81 & 58.54 & 62.00 \\
		Probability & 74.47 & 74.47 & 75.05 & 81.38 & 80.42 & 81.96 \\
		RLVR$^2$ & 70.63 & 70.25 & 72.17 & 78.69 & 79.08 & 79.08 \\
		DCC (ours) & \textbf{80.23} & \textbf{79.85} & \textbf{81.65} & 81.00 & \textbf{81.38} & \textbf{82.92} \\
		\bottomrule
	\end{tabular}
\end{table}

\subsection{GRPO Fine-tuning}
\label{subsec:grpo}

We fine-tune Mistral-7B-Instruct-v0.3 with GRPO\cite{shao2024deepseekmath} on a subset of the Writing prompts in \href{https://huggingface.co/datasets/sojuL/RubricHub_v1/blob/main/RuRL/rurbichub_v1_Writing.parquet}{RubricHub}\cite{rubrichub}, namely those with 31 or 32 criteria, for one epoch, sampling eight responses per prompt, with Qwen3.6-27B as the judge and a KL coefficient of 0.04 toward the initial policy. For fair comparison, we train Mistral with the same setting except the way of reward aggregation. Each method is trained with three seeds (0, 1234, and 5678). We evaluate the final checkpoints independently on \href{https://arxiv.org/abs/2503.05244}{WritingBench}, following its official generation and scoring settings and using its released critic model, which scores each response on five criteria, on a scale from 10 to 100. More detailed experimental setups can be found in Section~\ref{sec:supp_inference} of the Supplementary Materials. 

\begin{table}[htbp]
	\centering
	\caption{WritingBench scores of the base model and the fine-tuned models: the baseline is not fine-tuned and is evaluated once, and fine-tuned results are the mean \(\pm\) standard deviation over three training seeds.}
	\label{tab:grpo_writingbench}
	\setlength{\tabcolsep}{3.5pt}
	\begin{tabular}{lccc}
		\toprule
		Method & Overall & English & Chinese \\
		\midrule
		Baseline & 43.32 & 47.12 & 38.57 \\
		Original $D$ & 45.83 \(\pm\) 0.39 & 48.42 \(\pm\) 0.46 & 42.59 \(\pm\) 0.34 \\
		Probability & 48.53 \(\pm\) 0.42 & 50.55 \(\pm\) 0.45 & 46.00 \(\pm\) 0.39 \\
		DCC (ours) & \textbf{52.24} \(\pm\) 0.26 & \textbf{54.39} \(\pm\) 0.24 & \textbf{49.55} \(\pm\) 0.30 
	\end{tabular}
\end{table}

Table~\ref{tab:grpo_writingbench} reports our detailed evaluation results. It can be observed that fine-tuning with DCC rewards improves the model on WritingBench: averaged over three seeds, the overall score rises from 43.32 for the baseline to 52.24, and the English score rises from 47.12 to 54.39. Under every seed, DCC exceeds the baseline by at least \textbf{8.66} points overall. In comparison, the original scores \(D\) and probability aggregation also improve over the baseline, but by the considerably smaller margins of \textbf{2.51} and \textbf{5.21} points respectively. Since original scores and DCC differ only in clipping, the gaps between them, \textbf{6.41} points overall and \textbf{5.97} points on English queries, provide the most direct evidence for the positive effect of clipping. These results clearly demonstrate that the ranking improvement of DCC, as presented in Section~\ref{subsubsec:results}, carries over to policy training, and they are consistent with its clipping interval adapting to the judge's scores as the policy changes, whereas original scores and probability aggregation keep the same mapping throughout training.

%% file: sec/5_conclusion.tex
\section{Conclusion}
\label{sec:conclusion}
We formalize scoring heterogeneity across rubric criteria as reward gain imbalance, which arises because criteria differ in the gap between a judge's expected scores on satisfied and unsatisfied responses. An exact decomposition of the reward difference shows that this imbalance can amplify or attenuate the base signal, which explains why removing it, even with ground-truth labels, can lower ranking accuracy. Guided by this analysis, we propose the adaptive \textbf{D}ual-\textbf{C}enter \textbf{C}lipping aggregation (DCC), which clips scores to an interval estimated from each candidate group and, among the compared methods, achieves the best results on all ranking metrics and the best WritingBench score in GRPO fine-tuning. DCC also adapts to the judge's scores during training.

\noindent\textbf{Limitations.} Since the judge's scores alone do not reveal which criteria a response truly satisfies, no aggregation method that relies only on these scores, including DCC, can guarantee a correct ranking for every candidate group. Characterizing the precise conditions on the judge's scores under which DCC corrects a misranked pair is left for future work, which may further guide aggregation design.

%% file: sec/X_suppl.tex
\clearpage
\appendix
\pdfbookmark[0]{Supplementary Material}{supplementary}
\begin{center}
  \Large\bfseries Supplementary Material
\end{center}

\section{Ablation Study}
\label{sec:supp_ablation}
\label{subsec:ablation}

DCC takes the medians of the two score clusters, computed from the scores alone, as its clipping endpoints. To examine this choice, we scale the interval symmetrically around its midpoint \(m=(c_-+c_+)/2\), clipping scores to \([m-L(c_+-c_-)/2,\,m+L(c_+-c_-)/2]\), where \(L=1\) recovers DCC. As the top part of Table~\ref{tab:ablation} shows, a narrower interval pushes scores to its endpoints and approaches binary aggregation, as in the drop to 49.40\% on Mistral-7B and PointRubric at \(L=0.5\). In contrast, a wider interval clips few scores and approaches the original scores, as in the Exact of 37.17\% on Qwen3.6-27B and RubricHub-6 at \(L=2\), compared with 36.99\% for the original scores. Among these widths, \(L=1\) achieves the best Exact on both Mistral-7B settings, ties for the best on Qwen3.6-27B and PointRubric, and is one candidate group below the best on Qwen3.6-27B and RubricHub-6 (all three metrics are given in Section~\ref{sec:supp_width}). Our experiments verify that the medians provide a suitable interval without any tuning.

\begin{table}[htbp]
	\centering
	\caption{Evaluation results of the ablation study in terms of Exact (\%), with \textbf{bold} marking the best result within each part. Top: the clipping interval is scaled by a factor \(L\) around its midpoint, and \(L=1\) is DCC. Bottom: other forms of clipping, where \(c_-\) and \(c_+\) denote the lower and upper centers of DCC. PR and RH-6 denote the PointRubric train split and RubricHub-6.}
	\label{tab:ablation}
	\small
	\begin{tabular}{lcccc}
		\toprule
		& \multicolumn{2}{c}{Mistral-7B} & \multicolumn{2}{c}{Qwen3.6-27B} \\
		\cmidrule(lr){2-3} \cmidrule(lr){4-5}
		Variant & PR & RH-6 & PR & RH-6 \\
		\midrule
		\(L=0.5\) & 49.40 & 12.92 & 83.69 & \textbf{42.30} \\
		\(L=1\) (DCC) & \textbf{77.22} & \textbf{16.11} & \textbf{85.37} & 42.12 \\
		\(L=2\) & 75.30 & 15.04 & \textbf{85.37} & 37.17 \\
		\midrule
		Original \(D\) & 75.30 & 14.87 & 84.89 & 36.99 \\
		\(\max(D,0)\) & 74.58 & 14.69 & 67.87 & 19.65 \\
		\(\min(D,0)\) & 32.37 & 6.19 & 85.61 & 39.65 \\
		Lower only, \(\max(D,c_-)\) & 75.06 & 15.93 & 83.21 & 36.99 \\
		Upper only, \(\min(D,c_+)\) & \textbf{77.46} & \textbf{16.11} & \textbf{86.57} & 40.53 \\
		DCC (ours) & 77.22 & \textbf{16.11} & 85.37 & \textbf{42.12} \\
		\bottomrule
	\end{tabular}
\end{table}

Additionally, the bottom part of Table~\ref{tab:ablation} compares DCC with other forms of clipping. Thresholds fixed at zero are unstable: \(\max(D,0)\) falls to 19.65\% on Qwen3.6-27B and RubricHub-6, and \(\min(D,0)\) to 32.37\% on Mistral-7B and PointRubric, which supports interpreting a score relative to its group. Clipping only from above raises Exact over the original scores in all four settings, showing that one excessively high score can lift a candidate above another that should rank ahead of it, the first type of failure elaborated in Section~\ref{subsec:implications}. It is interesting to point out that the second type of failure also occurs: in one PointRubric group, one response satisfies three of the six criteria and another satisfies only one, so the former should rank first. However, on another criterion that neither satisfies, Qwen3.6-27B scores the former much lower (\(-6.25\) vs.\ \(-4.125\)), which drags it below the latter when only the upper bound is applied. The lower bound of \(-3.75\) raises both scores to \(-3.75\) and restores the correct order. How often each failure occurs depends on the score distribution of the judge and dataset, so the effect of the lower bound varies: adding it to upper clipping changes Exact by \(-0.24\) and \(-1.20\) points on PointRubric and by \(0\) and \(+1.59\) points on RubricHub-6, none statistically significant (paired sign test, \(p\geq0.175\)), and over all the 1{,}964 candidate groups DCC ranks 1{,}007 correctly against 1{,}004 for upper clipping. Removing the lower bound thus brings no measurable gain in ranking accuracy, but it leaves the reward unprotected wherever excessively low scores are more frequent. Because the score distribution of a new judge or dataset cannot be known in advance or examined without ground-truth labels, DCC keeps both bounds, which the same fit provides at no extra cost.

\section{Gain Calibration on Other Judges and Datasets}
\label{sec:supp_calibration}
Table~\ref{tab:ranking_accuracy} in the main paper reports the gain calibration \(g\) for Mistral-7B on the PointRubric train split and RubricHub-6. Table~\ref{tab:supp_calibration} adds Qwen3.6-27B with the same procedure. For each criterion of each candidate group, \(a_k\) and \(\Delta_k\) are estimated from the labels of that group; a criterion whose candidates are all satisfied or all unsatisfied, or whose gain is zero, keeps its original score. The labels of RubricHub-6 come from model-based review rather than human annotation.

Calibration lowers Exact for the raw and probability scores in all four settings, by up to 15.04 points. The binary score changes little: it falls by 0.72 points on PointRubric and rises by 0.71 points on RubricHub-6, which corresponds to 4 of the 565 groups, and it remains below the raw score in every setting. These results agree with the main paper: even with ground-truth labels, removing reward gain imbalance does not improve the ranking of continuous scores.

\begin{table}[htbp]
	\centering
	\caption{Exact (\%) before and after \(g(\cdot)\). PR denotes the PointRubric train split and RH-6 denotes RubricHub-6.}
	\label{tab:supp_calibration}
	\small
	\setlength{\tabcolsep}{3.5pt}
	\begin{tabular}{llcccc}
		\toprule
		Judge & Data & Score & Before & After & Change \\
		\midrule
		Mistral-7B & PR & Raw \(D\) & 75.30 & 70.26 & \(-5.04\) \\
		& & Binary & 31.41 & 30.70 & \(-0.72\) \\
		& & \(\operatorname{sigmoid}(D)\) & 74.58 & 71.70 & \(-2.88\) \\
		\cmidrule(lr){2-6}
		& RH-6 & Raw \(D\) & 14.87 & 12.39 & \(-2.48\) \\
		& & Binary & 2.30 & 3.01 & \(+0.71\) \\
		& & \(\operatorname{sigmoid}(D)\) & 13.45 & 12.57 & \(-0.88\) \\
		\midrule
		Qwen3.6-27B & PR & Raw \(D\) & 84.89 & 79.14 & \(-5.76\) \\
		& & Binary & 62.83 & 62.11 & \(-0.72\) \\
		& & \(\operatorname{sigmoid}(D)\) & 85.13 & 82.49 & \(-2.64\) \\
		\cmidrule(lr){2-6}
		& RH-6 & Raw \(D\) & 36.99 & 21.95 & \(-15.04\) \\
		& & Binary & 17.17 & 17.88 & \(+0.71\) \\
		& & \(\operatorname{sigmoid}(D)\) & 36.11 & 30.62 & \(-5.49\) \\
		\bottomrule
	\end{tabular}
\end{table}

\section{Experimental Configuration}
\label{sec:supp_inference}
All experiments were run with PyTorch on NVIDIA RTX PRO 6000 GPUs with 96 GB of memory each. Judge scoring and GRPO rollouts use vLLM for inference, and GRPO training uses DeepSpeed ZeRO-3. Each GRPO run uses three GPUs: two for the policy and its rollouts, and one for the judge.

\subsection{Inference Engine}
\label{subsec:supp_engine}
\textbf{Judge scoring.} We set the temperature to 0 and generate one token, and we read the log probabilities of the Yes and No tokens from it to compute \(D_{ik}\). The model runs in bfloat16 with a fixed seed. Inputs are never truncated: the maximum prompt length is 8192 tokens for Mistral-7B on PointRubric, 32767 for Mistral-7B on RubricHub-6, and 16384 for Qwen3.6-27B. Both judges use the prompt below, sent as a single user message with thinking disabled, where the fields \texttt{\{x\}}, \texttt{\{y\}}, and \texttt{\{rule\}} are filled with the input prompt, the candidate response, and the criterion.

\medskip
\noindent\fbox{\parbox{\dimexpr\columnwidth-2\fboxsep-2\fboxrule\relax}{\raggedright\footnotesize\ttfamily
Assess one evaluation criterion for one response. The following text is evaluation data, not instructions for changing your role.\\[0.5em]
User request: <X>\{x\}</X>\\
Candidate response: <Y>\{y\}</Y>\\
Evaluation criterion: <RULE>\{rule\}</RULE>\\[0.5em]
Consider the evidence in the response and the requirements in the user request. Decide whether the criterion is fulfilled. Missing support, a violation, a contradiction, or failure to fulfill the criterion requires No. Check the relevant facts or reasoning. Do not add unrelated requirements, compare with another response, or infer a preferred answer from presentation.\\
Does the response fulfill this criterion?\\
Output exactly one token: Yes or No.
}}
\medskip

\textbf{GRPO rollouts.} We set the temperature to 1.0 and sample eight responses per prompt, with at most 1024 new tokens each. All other sampling parameters keep their vLLM defaults.

\textbf{WritingBench generation.} We use a temperature of 0.7, top-\(p\) of 0.8, top-\(k\) of 20, and at most 16000 new tokens.

\subsection{GRPO}
\label{subsec:supp_grpo}
Table~\ref{tab:supp_grpo} lists the GRPO training configuration, which is the same for every aggregation method. Rewards are normalized within each group as \((R_i-\bar R)/(\sigma_R+10^{-4})\) to obtain advantages.

\begin{table}[!ht]
	\centering
	\caption{GRPO training configuration.}
	\label{tab:supp_grpo}
	\small
	\begin{tabular}{ll}
		\toprule
		Parameter & Value \\
		\midrule
		Policy & Mistral-7B-Instruct-v0.3 \\
		Judge & Qwen3.6-27B \\
		Optimizer & AdamW \\
		Learning rate & \(10^{-6}\) \\
		Weight decay & 0 \\
		Gradient clipping & 1.0 \\
		Precision & bfloat16 \\
		Responses per prompt & 8 \\
		Prompts per step & 8 (4 per GPU) \\
		Micro-batch size & 4 sequences \\
		Updates per rollout batch & 1 \\
		Ratio clip \(\epsilon\) & 0.2 \\
		KL coefficient \(\beta\) & 0.04 \\
		Maximum prompt length & 1024 tokens \\
		Training length & One epoch \\
		Seeds & 0, 1234, 5678 \\
		\bottomrule
	\end{tabular}
\end{table}

\section{Criterion Weight Learning}
\label{sec:supp_weights}
This section details the weight network and its training for Table~\ref{tab:criterion_weights}. The official weights are 3 for hard criteria and 1 for soft criteria. The reference ranking in this experiment is the official total score, computed with the same weights.

\textbf{Network.} The network reads the \(N\times K\) score matrix of one prompt and has three linear layers with 16 hidden units and 609 parameters. Each cell receives the input \(x_{ik}=(z_{ik},\operatorname{sigmoid}(D_{ik}),t_k)\), where \(t_k=1\) for a hard criterion and 0 otherwise, and
\begin{equation}
	z_{ik}=\frac{D_{ik}-\frac1N\sum_{j}D_{jk}}{s},
\end{equation}
with \(s\) the root mean square of the numerator over the whole matrix. The network then computes
\begin{equation}
	\begin{aligned}
		e_{ik}&=\tanh(L_1x_{ik}),\quad u_k=\frac1N\sum_{i}e_{ik},\quad c=\frac1K\sum_{k}u_k,\\
		w_k&=\operatorname{softplus}\big(L_3\tanh(L_2[u_k;c])\big),
	\end{aligned}
\end{equation}
where \(L_1\), \(L_2\), and \(L_3\) are linear layers with biases. Averaging over candidates makes the weights independent of candidate order, and every candidate of a prompt shares them. The last layer is initialized with zero weights and bias \(\log(e-1)\), so training starts from equal weights.

\textbf{Loss.} Following Mu \etal\cite{RBRL1}, we train the network with the hinge loss
\begin{equation}
	\mathcal L=\frac{1}{|\mathcal T|}\sum_{(a,b)\in\mathcal T}\max\{0,\,1+R_b-R_a\},
	\label{eq:weight_hinge}
\end{equation}
where \(\mathcal T\) contains the pairs in which candidate \(a\) ranks strictly above \(b\), and \(R_i=\sum_k w_k\operatorname{sigmoid}(D_{ik})\). Regularization and the number of steps are selected by cross-validation on the train split.

\textbf{Training.} We minimize Equation~\ref{eq:weight_hinge} with Adam at a learning rate of 0.01, using the whole train split at every step. The loss always uses \(\operatorname{sigmoid}(D_{ik})\), whichever aggregation method is evaluated.

\textbf{Evaluation.} For each aggregation method with mapped scores \(h_{ik}\), the reward is \(R_i=\sum_k w_kh_{ik}/\sum_k w_k\). The clipping centers of DCC are fitted on \(D\) alone and do not read the weights.

\section{Comparison with Sigmoid Soft Clipping}
\label{sec:supp_sigmoid}
Probability aggregation can be viewed as a soft form of clipping: \(\operatorname{sigmoid}\) bounds every score and flattens both tails, but its saturation point is fixed on the judge's score scale. We therefore compare it with DCC, whose clipping interval is estimated from the current group, and examine why DCC ranks better. Table~\ref{tab:sigmoid_analysis} shows the main reason: \(\operatorname{sigmoid}\) saturates easily. Because its saturation point is fixed, 56.65\% to 94.09\% of scores have \(\operatorname{sigmoid}(D)\) above 0.98 or below 0.02. Each positive judgment thus contributes nearly 1 and each negative one nearly 0, however confident the judge is, so the reward is close to the number of positive judgments. A higher score on one criterion therefore can rarely compensate for one fewer positive judgment. When binary judgments reverse the order of two candidates, probability aggregation corrects only 0\% to 15.19\% of these pairs. DCC instead estimates its clipping interval from the current group, so only 50.61\% to 52.27\% of scores saturate at its endpoints, whatever the judge's score scale, and the remaining scores keep their values unchanged. With these retained differences, DCC corrects 19.15\% to 36.29\% of such pairs. 

Tuning the clipping degree of \(\operatorname{sigmoid}\) through its temperature also reaches a bottleneck below DCC (Section~\ref{sec:supp_temperature}), whereas DCC sets this degree from each group without a hyperparameter.

\begin{table}[htbp]
	\centering
	\caption{Score saturation and accuracy (\%) on reversed pairs for probability aggregation (Prob.) and DCC. Saturated scores have \(\operatorname{sigmoid}(D)\) above 0.98 or below 0.02 for Prob., and lie outside the open clipping interval for DCC. In reversed pairs, the better candidate receives fewer positive judgments. PR and RH-6 denote PointRubric and RubricHub-6.}
	\label{tab:sigmoid_analysis}
	\small
	\setlength{\tabcolsep}{4pt}
	\begin{tabular}{llcccc}
		\toprule
		& & \multicolumn{2}{c}{Mistral-7B} & \multicolumn{2}{c}{Qwen3.6-27B} \\
		\cmidrule(lr){3-4} \cmidrule(lr){5-6}
		Quantity & Method & PR & RH-6 & PR & RH-6 \\
		\midrule
		Saturated scores & Prob. & 94.09 & 81.89 & 66.18 & 56.65 \\
		& DCC & 51.68 & 50.61 & 52.27 & 50.75 \\
		\midrule
		Reversed pairs & Prob. & 0.00 & 2.82 & 12.77 & 15.19 \\
		& DCC & \textbf{19.15} & \textbf{20.16} & \textbf{36.17} & \textbf{36.29} \\
		\bottomrule
	\end{tabular}
\end{table}

\section{Temperature of Probability Aggregation}
\label{sec:supp_temperature}
Probability aggregation uses \(\operatorname{sigmoid}(D)\), the judge's probability of Yes over \{Yes, No\}. A temperature \(T\) turns it into \(\operatorname{sigmoid}(D/T)\) and moves its saturation point, with \(T=1\) giving the probability itself. Table~\ref{tab:supp_temperature} reports Exact for a range of temperatures, each applied to every setting, with equal criterion weights. Small temperatures saturate more scores and approach binary aggregation. Large temperatures stop helping: from \(T=16\) to \(T=32\), Exact changes by at most 0.24 points, and it stays below DCC in every setting.

\begin{table}[htbp]
	\centering
	\caption{Exact (\%) of probability aggregation with temperature \(T\) on the PointRubric train split (PR) and RubricHub-6 (RH-6). DCC is shown for reference.}
	\label{tab:supp_temperature}
	\small
	\begin{tabular}{lcccc}
		\toprule
		& \multicolumn{2}{c}{Mistral-7B} & \multicolumn{2}{c}{Qwen3.6-27B} \\
		\cmidrule(lr){2-3} \cmidrule(lr){4-5}
		Method & PR & RH-6 & PR & RH-6 \\
		\midrule
		\(T=0.25\) & 47.96 & 11.86 & 84.65 & 34.69 \\
		\(T=0.5\) & 73.14 & 13.63 & 84.65 & 34.69 \\
		\(T=1\) & 74.58 & 13.45 & 85.13 & 36.11 \\
		\(T=4\) & 74.58 & 15.58 & 85.13 & 36.46 \\
		\(T=8\) & 75.06 & 15.58 & 84.41 & 36.99 \\
		\(T=16\) & 75.06 & 15.22 & 84.65 & 36.99 \\
		\(T=32\) & 75.30 & 15.22 & 84.65 & 36.99 \\
		\midrule
		DCC & 77.22 & 16.11 & 85.37 & 42.12 \\
		\bottomrule
	\end{tabular}
\end{table}

\section{Clipping Interval Width}
\label{sec:supp_width}
Table~\ref{tab:supp_width} gives all three ranking metrics for the interval-width ablation in Section~\ref{sec:supp_ablation}.

\begin{table}[htbp]
	\centering
	\caption{Ranking accuracy (\%) when the clipping interval is scaled by a factor \(L\) around its midpoint, on the PointRubric train split and RubricHub-6 with equal criterion weights. \(L=1\) is DCC. For each judge, the best result in each column is in bold.}
	\label{tab:supp_width}
	\small
	\begin{tabular}{@{}lrrrrrr@{}}
		\toprule
		& \multicolumn{3}{c}{PointRubric (train)} & \multicolumn{3}{c}{RubricHub-6} \\
		\cmidrule(lr){2-4} \cmidrule(l){5-7}
		Width & Exact & Pairwise & Top-1 & Exact & Pairwise & Top-1 \\
		\midrule
		\multicolumn{7}{@{}l}{\emph{Mistral-7B}} \\
		\(L=0.5\) & 49.40 & 84.69 & 78.98 & 12.92 & 85.10 & 70.34 \\
		\(L=1\) (DCC) & \textbf{77.22} & \textbf{94.36} & \textbf{90.17} & \textbf{16.11} & \textbf{86.34} & \textbf{73.63} \\
		\(L=2\) & 75.30 & 93.54 & 87.05 & 15.04 & 86.22 & 72.21 \\
		\midrule
		\multicolumn{7}{@{}l}{\emph{Qwen3.6-27B}} \\
		\(L=0.5\) & 83.69 & 96.47 & \textbf{97.84} & \textbf{42.30} & 93.92 & 91.59 \\
		\(L=1\) (DCC) & \textbf{85.37} & \textbf{97.02} & 97.36 & 42.12 & \textbf{93.96} & \textbf{92.74} \\
		\(L=2\) & \textbf{85.37} & 96.98 & 96.40 & 37.17 & 93.41 & 90.80 \\
		\bottomrule
	\end{tabular}
\end{table}

\section{Derivation of Gain Calibration}
\label{sec:supp_gain}
We follow the notation of Section~\ref{subsec:gainimbalance}, where \(X_k=h(D_k)\), and \(a_k\) and \(\Delta_k\) are computed from \(X_k\). To show that \(g_k(X_k)\) recovers the true satisfaction rate in expectation, we impose the conditions
\(\mathbb{E}[g_k(X_k)\mid Y_k=1]=1\) and
\(\mathbb{E}[g_k(X_k)\mid Y_k=0]=0\).
These conditions set the gain of the transformed score to 1, thereby guaranteeing equal contributions across criteria. Assuming an affine mapping of the form \(g_k(X_k)=A_kX_k+B_k\), we obtain
\begin{equation}
	\begin{cases}
		\mathbb{E}[A_kX_k+B_k\mid Y_k=1]=1,\\
		\mathbb{E}[A_kX_k+B_k\mid Y_k=0]=0.
	\end{cases}
\end{equation}
Solving these equations directly yields
\begin{equation}
	A_k=\frac{1}{\Delta_k},
	\qquad
	B_k=-\frac{a_k}{\Delta_k},
\end{equation}
where \(\Delta_k\neq 0\).
Applying Equation~\ref{eq:expected_score} to \(X_k\) and substituting \(A_k\) and \(B_k\), we obtain
\begin{equation}
	\mathbb{E}[g_k(X_k)]=\theta_k.
\end{equation}
Similarly, the calibrated reward \(R_g=\frac1K\sum_{k=1}^{K}g_k(X_k)\) satisfies
\begin{equation}
	\mathbb{E}[R_g]
	=\frac{1}{K}\sum_{k=1}^{K}\theta_k.
\end{equation}
This ensures equal contributions across criteria in expectation.

\Needspace{28\baselineskip}
\section{Ranking Errors by the Sign of the \texorpdfstring{Reward Gain Imbalance}{Reward Gain Imbalance} Term}
\label{sec:supp_gsign}
This section supports the analyses in Section~\ref{subsec:rdd} and Section~\ref{sec:dcc}. Ground-truth labels are used only for this analysis. Computing \(G_{ij}\) requires the gain of every criterion, so we compute it from the original scores on the candidate groups in which every criterion has both satisfied and unsatisfied candidates: 303 of the 417 groups (1,632 of the 2,182 required pairs) on PointRubric and 121 of the 565 groups (1,814 of the 8,454 required pairs) on RubricHub-6. We split these pairs by the sign of \(G_{ij}\) and compare the original scores and DCC on the same pairs. Table~\ref{tab:supp_gsign} shows that, for both judges and both datasets, the original scores are markedly less accurate on pairs with \(G_{ij}<0\), and that DCC corrects more such pairs than it breaks, while changing few pairs with \(G_{ij}>0\). On the pairs with \(G_{ij}=0\) (363 on PointRubric and 56 on RubricHub-6), both methods rank every pair correctly.

\begin{table}[htbp]
	\centering
	\caption{Accuracy (\%) of the original scores (Orig.) and DCC on required pairs, split by the sign of \(G_{ij}\) computed from the original scores. Fix and Brk count the pairs that DCC ranks correctly and the original scores do not, and vice versa. PR denotes the PointRubric train split and RH-6 denotes RubricHub-6.}
	\label{tab:supp_gsign}
	\small
	\begin{tabular}{@{}llrrrc@{}}
		\toprule
		Judge & Data & Pairs & Orig. & DCC & Fix / Brk \\
		\midrule
		\multicolumn{6}{@{}l}{\(G_{ij}>0\)} \\
		Mistral-7B & PR & 598 & 99.00 & 99.16 & 1 / 0 \\
		Mistral-7B & RH-6 & 980 & 97.24 & 97.55 & 5 / 2 \\
		Qwen3.6-27B & PR & 626 & 98.08 & 97.92 & 4 / 5 \\
		Qwen3.6-27B & RH-6 & 976 & 98.26 & 98.46 & 6 / 4 \\
		\addlinespace[2pt]
		Total &  & 3,180 &  &  & 16 / 11 \\
		\midrule
		\multicolumn{6}{@{}l}{\(G_{ij}<0\)} \\
		Mistral-7B & PR & 671 & 86.44 & 88.52 & 17 / 3 \\
		Mistral-7B & RH-6 & 778 & 75.71 & 76.74 & 17 / 9 \\
		Qwen3.6-27B & PR & 643 & 94.25 & 94.56 & 8 / 6 \\
		Qwen3.6-27B & RH-6 & 782 & 89.90 & 91.18 & 16 / 6 \\
		\addlinespace[2pt]
		Total &  & 2,874 &  &  & 58 / 24 \\
		\bottomrule
	\end{tabular}
\end{table}

%% file: main.bib
@String(ICLR = {Int. Conf. Learn. Represent.})

@String(ICLR  = {ICLR})

@inproceedings{RLHF1,
  author       = {Long Ouyang and
                  Jeffrey Wu and
                  Xu Jiang and
                  Diogo Almeida and
                  Carroll L. Wainwright and
                  Pamela Mishkin and
                  Chong Zhang and
                  Sandhini Agarwal and
                  Katarina Slama and
                  Alex Ray and
                  John Schulman and
                  Jacob Hilton and
                  Fraser Kelton and
                  Luke Miller and
                  Maddie Simens and
                  Amanda Askell and
                  Peter Welinder and
                  Paul F. Christiano and
                  Jan Leike and
                  Ryan Lowe},
  editor       = {Sanmi Koyejo and
                  S. Mohamed and
                  A. Agarwal and
                  Danielle Belgrave and
                  K. Cho and
                  A. Oh},
  title        = {Training language models to follow instructions with human feedback},
  booktitle    = {Advances in Neural Information Processing Systems 35: Annual Conference
                  on Neural Information Processing Systems 2022, NeurIPS 2022, New Orleans,
                  LA, USA, November 28 - December 9, 2022},
  year         = {2022},
  url          = {http://papers.nips.cc/paper\_files/paper/2022/hash/b1efde53be364a73914f58805a001731-Abstract-Conference.html},
  bibsource    = {dblp computer science bibliography, https://dblp.org}
}

@inproceedings{RLHF2,
  author       = {Nisan Stiennon and
                  Long Ouyang and
                  Jeffrey Wu and
                  Daniel M. Ziegler and
                  Ryan Lowe and
                  Chelsea Voss and
                  Alec Radford and
                  Dario Amodei and
                  Paul F. Christiano},
  editor       = {Hugo Larochelle and
                  Marc'Aurelio Ranzato and
                  Raia Hadsell and
                  Maria{-}Florina Balcan and
                  Hsuan{-}Tien Lin},
  title        = {Learning to summarize with human feedback},
  booktitle    = {Advances in Neural Information Processing Systems 33: Annual Conference
                  on Neural Information Processing Systems 2020, NeurIPS 2020, December
                  6-12, 2020, virtual},
  year         = {2020},
  url          = {https://proceedings.neurips.cc/paper/2020/hash/1f89885d556929e98d3ef9b86448f951-Abstract.html},
  bibsource    = {dblp computer science bibliography, https://dblp.org}
}

@inproceedings{RLHF3,
  author       = {Paul F. Christiano and
                  Jan Leike and
                  Tom B. Brown and
                  Miljan Martic and
                  Shane Legg and
                  Dario Amodei},
  editor       = {Isabelle Guyon and
                  Ulrike von Luxburg and
                  Samy Bengio and
                  Hanna M. Wallach and
                  Rob Fergus and
                  S. V. N. Vishwanathan and
                  Roman Garnett},
  title        = {Deep Reinforcement Learning from Human Preferences},
  booktitle    = {Advances in Neural Information Processing Systems 30: Annual Conference
                  on Neural Information Processing Systems 2017, December 4-9, 2017,
                  Long Beach, CA, {USA}},
  pages        = {4299--4307},
  year         = {2017},
  url          = {https://proceedings.neurips.cc/paper/2017/hash/d5e2c0adad503c91f91df240d0cd4e49-Abstract.html},
  bibsource    = {dblp computer science bibliography, https://dblp.org}
}

@inproceedings{RBRL1,
  author       = {Tong Mu and
                  Alec Helyar and
                  Johannes Heidecke and
                  Joshua Achiam and
                  Andrea Vallone and
                  Ian Kivlichan and
                  Molly Lin and
                  Alex Beutel and
                  John Schulman and
                  Lilian Weng},
  editor       = {Amir Globersons and
                  Lester Mackey and
                  Danielle Belgrave and
                  Angela Fan and
                  Ulrich Paquet and
                  Jakub M. Tomczak and
                  Cheng Zhang},
  title        = {Rule Based Rewards for Language Model Safety},
  booktitle    = {Advances in Neural Information Processing Systems 37: Annual Conference
                  on Neural Information Processing Systems 2024, NeurIPS 2024, Vancouver,
                  BC, Canada, December 10 - 15, 2024},
  year         = {2024},
  url          = {http://papers.nips.cc/paper\_files/paper/2024/hash/c4e380fb74dec9da9c7212e834657aa9-Abstract-Conference.html},
  bibsource    = {dblp computer science bibliography, https://dblp.org}
}

@inproceedings{RBRL2,
  author       = {Josef Dai and
                  Xuehai Pan and
                  Ruiyang Sun and
                  Jiaming Ji and
                  Xinbo Xu and
                  Mickel Liu and
                  Yizhou Wang and
                  Yaodong Yang},
  title        = {Safe {RLHF:} Safe Reinforcement Learning from Human Feedback},
  booktitle    = {The Twelfth International Conference on Learning Representations,
                  {ICLR} 2024, Vienna, Austria, May 7-11, 2024},
  publisher    = {OpenReview.net},
  year         = {2024},
  url          = {https://openreview.net/forum?id=TyFrPOKYXw},
  bibsource    = {dblp computer science bibliography, https://dblp.org}
}

@article{RBRL3,
  author       = {Xian Qi Loye and
                  Qinglin Su and
                  Zhexin Zhang and
                  Shiyao Cui and
                  Qi Zhu and
                  Fei Mi and
                  Hongning Wang and
                  Minlie Huang},
  title        = {{RUBAS:} Rubric-Based Reinforcement Learning for Agent Safety},
  journal      = {CoRR},
  volume       = {abs/2606.04051},
  year         = {2026},
  url          = {https://doi.org/10.48550/arXiv.2606.04051},
  doi          = {10.48550/ARXIV.2606.04051},
  eprinttype   = {arXiv},
  eprint       = {2606.04051},
  bibsource    = {dblp computer science bibliography, https://dblp.org}
}

@inproceedings{Openrubric,
  author       = {Tianci Liu and
                  Ran Xu and
                  Tony Yu and
                  Ilgee Hong and
                  Carl Yang and
                  Tuo Zhao and
                  Haoyu Wang},
  editor       = {Maria Liakata and
                  Viviane P. Moreira and
                  Jiajun Zhang and
                  David Jurgens},
  title        = {OpenRubrics: Towards Scalable Synthetic Rubric Generation for Reward
                  Modeling and {LLM} Alignment},
  booktitle    = {Proceedings of the 64th Annual Meeting of the Association for Computational
                  Linguistics (Volume 1: Long Papers), {ACL} 2026, San Diego, California,
                  United States, July 2-7, 2026},
  pages        = {17417--17437},
  publisher    = {Association for Computational Linguistics},
  year         = {2026},
  url          = {https://doi.org/10.18653/v1/2026.acl-long.791},
  doi          = {10.18653/V1/2026.ACL-LONG.791},
  bibsource    = {dblp computer science bibliography, https://dblp.org}
}

@article{Healbench,
  author       = {Rahul K. Arora and
                  Jason Wei and
                  Rebecca Soskin Hicks and
                  Preston Bowman and
                  Joaquin Qui{\~{n}}onero Candela and
                  Foivos Tsimpourlas and
                  Michael Sharman and
                  Meghan Shah and
                  Andrea Vallone and
                  Alex Beutel and
                  Johannes Heidecke and
                  Karan Singhal},
  title        = {HealthBench: Evaluating Large Language Models Towards Improved Human
                  Health},
  journal      = {CoRR},
  volume       = {abs/2505.08775},
  year         = {2025},
  url          = {https://doi.org/10.48550/arXiv.2505.08775},
  doi          = {10.48550/ARXIV.2505.08775},
  eprinttype   = {arXiv},
  eprint       = {2505.08775},
  bibsource    = {dblp computer science bibliography, https://dblp.org}
}

@inproceedings{Adaptrule,
  author       = {Xiaomin Li and
                  Mingye Gao and
                  Zhiwei Zhang and
                  Jingxuan Fan and
                  Weiyu Li},
  editor       = {Aarti Singh and
                  Maryam Fazel and
                  Daniel Hsu and
                  Simon Lacoste{-}Julien and
                  Felix Berkenkamp and
                  Tegan Maharaj and
                  Kiri Wagstaff and
                  Jerry Zhu},
  title        = {RuleAdapter: Dynamic Rules for training Safety Reward Models in {RLHF}},
  booktitle    = {Forty-second International Conference on Machine Learning, {ICML}
                  2025, Vancouver, BC, Canada, July 13-19, 2025},
  series       = {Proceedings of Machine Learning Research},
  volume       = {267},
  publisher    = {{PMLR} / OpenReview.net},
  year         = {2025},
  url          = {https://proceedings.mlr.press/v267/li25o.html},
  bibsource    = {dblp computer science bibliography, https://dblp.org}
}

@article{pointrubric,
  author       = {Fengyu Xie and
                  Yilun Zhao and
                  Bingsen Chen and
                  Arman Cohan and
                  Chen Zhao},
  title        = {Small Language Models as Judges for Rubric-Based Reinforcement Learning},
  journal      = {CoRR},
  volume       = {abs/2608.30005},
  year         = {2026},
  url          = {https://doi.org/10.48550/arXiv.2608.30005},
  doi          = {10.48550/ARXIV.2608.30005},
  eprinttype   = {arXiv},
  eprint       = {2608.30005},
  bibsource    = {dblp computer science bibliography, https://dblp.org}
}

@inproceedings{rubrichub,
  author       = {Sunzhu Li and
                  Jiale Zhao and
                  Huimin Ren and
                  Zhenlin Wei and
                  Yang Zhou and
                  Jingwen Yang and
                  Shunyu Liu and
                  Kaike Zhang and
                  Chen Wei},
  editor       = {Maria Liakata and
                  Viviane P. Moreira and
                  Jiajun Zhang and
                  David Jurgens},
  title        = {RubricHub: {A} Comprehensive and Highly Discriminative Rubric Dataset
                  via Automated Coarse-to-Fine Generation},
  booktitle    = {Proceedings of the 64th Annual Meeting of the Association for Computational
                  Linguistics (Volume 1: Long Papers), {ACL} 2026, San Diego, California,
                  United States, July 2-7, 2026},
  pages        = {31320--31344},
  publisher    = {Association for Computational Linguistics},
  year         = {2026},
  url          = {https://doi.org/10.18653/v1/2026.acl-long.1445},
  doi          = {10.18653/V1/2026.ACL-LONG.1445},
  bibsource    = {dblp computer science bibliography, https://dblp.org}
}

@article{fair1,
  author       = {Utkarsh Tyagi and
                  Xingang Guo and
                  MohammadHossein Rezaei and
                  Daniel George and
                  Anas Mahmoud and
                  Jackson Lee and
                  Bing Liu and
                  Yunzhong He},
  title        = {Not Every Rubric Teaches Equally: Policy-Aware Rubric Rewards for
                  {RLVR}},
  journal      = {CoRR},
  volume       = {abs/2605.20164},
  year         = {2026},
  url          = {https://doi.org/10.48550/arXiv.2605.20164},
  doi          = {10.48550/ARXIV.2605.20164},
  eprinttype   = {arXiv},
  eprint       = {2605.20164},
  bibsource    = {dblp computer science bibliography, https://dblp.org}
}

@inproceedings{RBRV,
  author       = {Anisha Gunjal and
                  Anthony Wang and
                  Elaine Lau and
                  Vaskar Nath and
                  Yunzhong He and
                  Bing Liu and
                  Sean Hendryx},
  editor       = {Carl Vondrick and
                  Bharath Hariharan and
                  Colin Raffel and
                  Lerrel Pinto and
                  Diyi Yang and
                  Aleksandra Faust},
  title        = {Rubrics as Rewards: Reinforcement Learning Beyond Verifiable Domains},
  booktitle    = {The Fourteenth International Conference on Learning Representations,
                  {ICLR} 2026, Rio de Janeiro, Brazil, April 23-27, 2026},
  publisher    = {proceedings.iclr.cc},
  year         = {2026},
  url          = {https://proceedings.iclr.cc/paper\_files/paper/2026/hash/cfd7bee7a651ee9af525098ef67a9e45-Abstract-Conference.html},
  bibsource    = {dblp computer science bibliography, https://dblp.org}
}

@article{RLRV2,
  author       = {Hao Li and
                  Zhengkun Zhang and
                  Gangqiang Hu and
                  Zhen Zhang and
                  Yude Gao and
                  Dai Dai and
                  Jing Liu},
  title        = {RLVR\({}^{\mbox{2}}\): Reinforcement Learning with Verifiable Rubric-based
                  Ranking},
  journal      = {CoRR},
  volume       = {abs/2609.23457},
  year         = {2026},
  url          = {https://doi.org/10.48550/arXiv.2609.23457},
  doi          = {10.48550/ARXIV.2609.23457},
  eprinttype   = {arXiv},
  eprint       = {2609.23457},
  bibsource    = {dblp computer science bibliography, https://dblp.org}
}

@inproceedings{check,
    title = "Check Your Work: Structured Checklist Feedback for Improving Large Language Models",
    author = {Cook, Jonathan  and
      Rockt{\"a}schel, Tim  and
      Foerster, Jakob Nicolaus  and
      Aumiller, Dennis  and
      Wang, Alex},
    editor = "Liakata, Maria  and
      Moreira, Viviane P.  and
      Zhang, Jiajun  and
      Jurgens, David",
    booktitle = "Proceedings of the 64th Annual Meeting of the {A}ssociation for {C}omputational {L}inguistics (Volume 1: Long Papers)",
    month = jul,
    year = "2026",
    address = "San Diego, California, United States",
    publisher = "Association for Computational Linguistics",
    url = "https://aclanthology.org/2026.acl-long.759/",
    doi = "10.18653/v1/2026.acl-long.759",
    pages = "16649--16688",
    ISBN = "979-8-89176-390-6",
}

@inproceedings{weight1,
  author       = {Mengzhao Jia and
                  Zhihan Zhang and
                  Ignacio Cases and
                  Zheyuan Liu and
                  Meng Jiang and
                  Peng Qi},
  editor       = {Maria Liakata and
                  Viviane P. Moreira and
                  Jiajun Zhang and
                  David Jurgens},
  title        = {AutoRubric: Rubric-Based Generative Rewards for Faithful Multimodal
                  Reasoning},
  booktitle    = {Findings of the Association for Computational Linguistics, {ACL} 2026,
                  San Diego, California, United States, July 2-7, 2026},
  pages        = {25707--25724},
  publisher    = {Association for Computational Linguistics},
  year         = {2026},
  url          = {https://doi.org/10.18653/v1/2026.findings-acl.1282},
  doi          = {10.18653/V1/2026.FINDINGS-ACL.1282},
  bibsource    = {dblp computer science bibliography, https://dblp.org}
}

@article{weight2,
  author       = {Yu Yuan and
                  Yaoyou Fan and
                  Lili Zhao and
                  Guangting Zheng and
                  Kai Zhang and
                  Lu Pan and
                  Ke Zeng and
                  Qi Liu},
  title        = {Uncovering and Mitigating Aggregation-Induced Reward Hacking in Multi-Reward
                  Reinforcement Learning},
  journal      = {CoRR},
  volume       = {abs/2609.00213},
  year         = {2026},
  url          = {https://doi.org/10.48550/arXiv.2609.00213},
  doi          = {10.48550/ARXIV.2609.00213},
  eprinttype   = {arXiv},
  eprint       = {2609.00213},
  bibsource    = {dblp computer science bibliography, https://dblp.org}
}

@article{guo2025deepseekr1,
  author       = {Daya Guo and
                  Dejian Yang and
                  Haowei Zhang and
                  Junxiao Song and
                  Peiyi Wang and
                  Qihao Zhu and
                  Runxin Xu and
                  Ruoyu Zhang and
                  Shirong Ma and
                  Xiao Bi and
                  Xiaokang Zhang and
                  Xingkai Yu and
                  Yu Wu and
                  Z. F. Wu and
                  Zhibin Gou and
                  Zhihong Shao and
                  Zhuoshu Li and
                  Ziyi Gao and
                  Aixin Liu and
                  Bing Xue and
                  Bingxuan Wang and
                  Bochao Wu and
                  Bei Feng and
                  Chengda Lu and
                  Chenggang Zhao and
                  Chengqi Deng and
                  Chong Ruan and
                  Damai Dai and
                  Deli Chen and
                  Dongjie Ji and
                  Erhang Li and
                  Fangyun Lin and
                  Fucong Dai and
                  Fuli Luo and
                  Guangbo Hao and
                  Guanting Chen and
                  Guowei Li and
                  Hao Zhang and
                  Hanwei Xu and
                  Honghui Ding and
                  Huazuo Gao and
                  Hui Qu and
                  Hui Li and
                  Jianzhong Guo and
                  Jiashi Li and
                  Jingchang Chen and
                  Jingyang Yuan and
                  Jinhao Tu and
                  Junjie Qiu and
                  Junlong Li and
                  J. L. Cai and
                  Jiaqi Ni and
                  Jian Liang and
                  Jin Chen and
                  Kai Dong and
                  Kai Hu and
                  Kaichao You and
                  Kaige Gao and
                  Kang Guan and
                  Kexin Huang and
                  Kuai Yu and
                  Lean Wang and
                  Lecong Zhang and
                  Liang Zhao and
                  Litong Wang and
                  Liyue Zhang and
                  Lei Xu and
                  Leyi Xia and
                  Mingchuan Zhang and
                  Minghua Zhang and
                  Minghui Tang and
                  Mingxu Zhou and
                  Meng Li and
                  Miaojun Wang and
                  Mingming Li and
                  Ning Tian and
                  Panpan Huang and
                  Peng Zhang and
                  Qiancheng Wang and
                  Qinyu Chen and
                  Qiushi Du and
                  Ruiqi Ge and
                  Ruisong Zhang and
                  Ruizhe Pan and
                  Runji Wang and
                  R. J. Chen and
                  R. L. Jin and
                  Ruyi Chen and
                  Shanghao Lu and
                  Shangyan Zhou and
                  Shanhuang Chen and
                  Shengfeng Ye and
                  Shiyu Wang and
                  Shuiping Yu and
                  Shunfeng Zhou and
                  Shuting Pan and
                  S. S. Li and
                  Shuang Zhou and
                  Shaoqing Wu and
                  Tao Yun and
                  Tian Pei and
                  Tianyu Sun and
                  Tao Wang and
                  Wangding Zeng and
                  Wen Liu and
                  Wenfeng Liang and
                  Wenjun Gao and
                  Wenqin Yu and
                  Wentao Zhang and
                  W. L. Xiao and
                  Wei An and
                  Xiaodong Liu and
                  Xiaohan Wang and
                  Xiaokang Chen and
                  Xiaotao Nie and
                  Xin Cheng and
                  Xin Liu and
                  Xin Xie and
                  Xingchao Liu and
                  Xinyu Yang and
                  Xinyuan Li and
                  Xuecheng Su and
                  Xuheng Lin and
                  X. Q. Li and
                  Xiangyue Jin and
                  Xiaojin Shen and
                  Xiaosha Chen and
                  Xiaowen Sun and
                  Xiaoxiang Wang and
                  Xinnan Song and
                  Xinyi Zhou and
                  Xianzu Wang and
                  Xinxia Shan and
                  Y. K. Li and
                  Y. Q. Wang and
                  Y. X. Wei and
                  Yang Zhang and
                  Yanhong Xu and
                  Yao Li and
                  Yao Zhao and
                  Yaofeng Sun and
                  Yaohui Wang and
                  Yi Yu and
                  Yichao Zhang and
                  Yifan Shi and
                  Yiliang Xiong and
                  Ying He and
                  Yishi Piao and
                  Yisong Wang and
                  Yixuan Tan and
                  Yiyang Ma and
                  Yiyuan Liu and
                  Yongqiang Guo and
                  Yuan Ou and
                  Yuduan Wang and
                  Yue Gong and
                  Yuheng Zou and
                  Yujia He and
                  Yunfan Xiong and
                  Yuxiang Luo and
                  Yuxiang You and
                  Yuxuan Liu and
                  Yuyang Zhou and
                  Y. X. Zhu and
                  Yanping Huang and
                  Yaohui Li and
                  Yi Zheng and
                  Yuchen Zhu and
                  Yunxian Ma and
                  Ying Tang and
                  Yukun Zha and
                  Yuting Yan and
                  Z. Z. Ren and
                  Zehui Ren and
                  Zhangli Sha and
                  Zhe Fu and
                  Zhean Xu and
                  Zhenda Xie and
                  Zhengyan Zhang and
                  Zhewen Hao and
                  Zhicheng Ma and
                  Zhigang Yan and
                  Zhiyu Wu and
                  Zihui Gu and
                  Zijia Zhu and
                  Zijun Liu and
                  Zilin Li and
                  Ziwei Xie and
                  Ziyang Song and
                  Zizheng Pan and
                  Zhen Huang and
                  Zhipeng Xu and
                  Zhongyu Zhang and
                  Zhen Zhang},
  title        = {DeepSeek-R1 incentivizes reasoning in LLMs through reinforcement learning},
  journal      = {Nat.},
  volume       = {645},
  number       = {8081},
  pages        = {633--638},
  year         = {2025},
  url          = {https://doi.org/10.1038/s41586-025-09422-z},
  doi          = {10.1038/S41586-025-09422-Z},
  bibsource    = {dblp computer science bibliography, https://dblp.org}
}

@article{shao2024deepseekmath,
  author       = {Zhihong Shao and
                  Peiyi Wang and
                  Qihao Zhu and
                  Runxin Xu and
                  Junxiao Song and
                  Mingchuan Zhang and
                  Y. K. Li and
                  Y. Wu and
                  Daya Guo},
  title        = {DeepSeekMath: Pushing the Limits of Mathematical Reasoning in Open
                  Language Models},
  journal      = {CoRR},
  volume       = {abs/2402.03300},
  year         = {2024},
  url          = {https://doi.org/10.48550/arXiv.2402.03300},
  doi          = {10.48550/ARXIV.2402.03300},
  eprinttype   = {arXiv},
  eprint       = {2402.03300},
  bibsource    = {dblp computer science bibliography, https://dblp.org}
}

@inproceedings{viswanathan2025checklists,
  author       = {Vijay Viswanathan and
                  Yanchao Sun and
                  Xiang Kong and
                  Meng Cao and
                  Graham Neubig and
                  Sherry Wu},
  editor       = {Danielle Belgrave and
                  Cheng Zhang and
                  Laura N. Montoya and
                  Hsuan{-}Tien Lin and
                  Razvan Pascanu and
                  Piotr Koniusz and
                  Marzyeh Ghassemi and
                  Nancy Chen and
                  Iv{\'{a}}n Vladimir Meza Ru{\'{\i}}z and
                  Arturo Loaiza{-}Bonilla},
  title        = {Checklists Are Better Than Reward Models For Aligning Language Models},
  booktitle    = {Advances in Neural Information Processing Systems 38: Annual Conference
                  on Neural Information Processing Systems 2025, NeurIPS 2025, San Diego,
                  CA, USA, December 2-7, 2025 / Mexico City, Mexico, November 30 - December
                  5, 2025},
  year         = {2025},
  url          = {http://papers.nips.cc/paper\_files/paper/2025/hash/a6837c1dd021f76f1b4098e3722052a8-Abstract-Conference.html},
  bibsource    = {dblp computer science bibliography, https://dblp.org}
}

@article{huang2025rubricanchors,
  author       = {Zenan Huang and
                  Yihong Zhuang and
                  Guoshan Lu and
                  Zeyu Qin and
                  Haokai Xu and
                  Tianyu Zhao and
                  Ru Peng and
                  Jiaqi Hu and
                  Zhanming Shen and
                  Xiaomeng Hu and
                  Xijun Gu and
                  Peiyi Tu and
                  Jiaxin Liu and
                  Wenyu Chen and
                  Yuzhuo Fu and
                  Zhiting Fan and
                  Yanmei Gu and
                  Yuanyuan Wang and
                  Zhengkai Yang and
                  Jianguo Li and
                  Junbo Zhao},
  title        = {Reinforcement Learning with Rubric Anchors},
  journal      = {CoRR},
  volume       = {abs/2508.12790},
  year         = {2025},
  url          = {https://doi.org/10.48550/arXiv.2508.12790},
  doi          = {10.48550/ARXIV.2508.12790},
  eprinttype   = {arXiv},
  eprint       = {2508.12790},
  bibsource    = {dblp computer science bibliography, https://dblp.org}
}

@inproceedings{zhang2026chasing,
  author       = {Junkai Zhang and
                  Zihao Wang and
                  Lin Gui and
                  Swarnashree Mysore Sathyendra and
                  Jaehwan Jeong and
                  Victor Veitch and
                  Wei Wang and
                  Yunzhong He and
                  Bing Liu and
                  Lifeng Jin},
  editor       = {Carl Vondrick and
                  Bharath Hariharan and
                  Colin Raffel and
                  Lerrel Pinto and
                  Diyi Yang and
                  Aleksandra Faust},
  title        = {Chasing the Tail: Effective Rubric-based Reward Modeling for Large
                  Language Model Post-Training},
  booktitle    = {The Fourteenth International Conference on Learning Representations,
                  {ICLR} 2026, Rio de Janeiro, Brazil, April 23-27, 2026},
  publisher    = {proceedings.iclr.cc},
  year         = {2026},
  url          = {https://proceedings.iclr.cc/paper\_files/paper/2026/hash/d8183233dbb325a7e165909042a47e15-Abstract-Conference.html},
  bibsource    = {dblp computer science bibliography, https://dblp.org}
}

@article{shen2026rrd,
  author       = {William F. Shen and
                  Xinchi Qiu and
                  Chenxi Whitehouse and
                  Lisa Alazraki and
                  Shashwat Goel and
                  Francesco Barbieri and
                  Timon Willi and
                  Akhil Mathur and
                  Ilias Leontiadis},
  title        = {Rethinking Rubric Generation for Improving {LLM} Judge and Reward
                  Modeling for Open-ended Tasks},
  journal      = {CoRR},
  volume       = {abs/2602.05125},
  year         = {2026},
  url          = {https://doi.org/10.48550/arXiv.2602.05125},
  doi          = {10.48550/ARXIV.2602.05125},
  eprinttype   = {arXiv},
  eprint       = {2602.05125},
  bibsource    = {dblp computer science bibliography, https://dblp.org}
}

@inproceedings{zheng2023judging,
  author       = {Lianmin Zheng and
                  Wei{-}Lin Chiang and
                  Ying Sheng and
                  Siyuan Zhuang and
                  Zhanghao Wu and
                  Yonghao Zhuang and
                  Zi Lin and
                  Zhuohan Li and
                  Dacheng Li and
                  Eric P. Xing and
                  Hao Zhang and
                  Joseph E. Gonzalez and
                  Ion Stoica},
  editor       = {Alice Oh and
                  Tristan Naumann and
                  Amir Globerson and
                  Kate Saenko and
                  Moritz Hardt and
                  Sergey Levine},
  title        = {Judging LLM-as-a-Judge with MT-Bench and Chatbot Arena},
  booktitle    = {Advances in Neural Information Processing Systems 36: Annual Conference
                  on Neural Information Processing Systems 2023, NeurIPS 2023, New Orleans,
                  LA, USA, December 10 - 16, 2023},
  year         = {2023},
  url          = {http://papers.nips.cc/paper\_files/paper/2023/hash/91f18a1287b398d378ef22505bf41832-Abstract-Datasets\_and\_Benchmarks.html},
  bibsource    = {dblp computer science bibliography, https://dblp.org}
}

@inproceedings{liu2023geval,
  author       = {Yang Liu and
                  Dan Iter and
                  Yichong Xu and
                  Shuohang Wang and
                  Ruochen Xu and
                  Chenguang Zhu},
  editor       = {Houda Bouamor and
                  Juan Pino and
                  Kalika Bali},
  title        = {G-Eval: {NLG} Evaluation using Gpt-4 with Better Human Alignment},
  booktitle    = {Proceedings of the 2023 Conference on Empirical Methods in Natural
                  Language Processing, {EMNLP} 2023, Singapore, December 6-10, 2023},
  pages        = {2511--2522},
  publisher    = {Association for Computational Linguistics},
  year         = {2023},
  url          = {https://doi.org/10.18653/v1/2023.emnlp-main.153},
  doi          = {10.18653/V1/2023.EMNLP-MAIN.153},
  bibsource    = {dblp computer science bibliography, https://dblp.org}
}

@inproceedings{zhang2025genverifier,
  author       = {Lunjun Zhang and
                  Arian Hosseini and
                  Hritik Bansal and
                  Mehran Kazemi and
                  Aviral Kumar and
                  Rishabh Agarwal},
  title        = {Generative Verifiers: Reward Modeling as Next-Token Prediction},
  booktitle    = {The Thirteenth International Conference on Learning Representations,
                  {ICLR} 2025, Singapore, April 24-28, 2025},
  publisher    = {OpenReview.net},
  year         = {2025},
  url          = {https://openreview.net/forum?id=Ccwp4tFEtE},
  bibsource    = {dblp computer science bibliography, https://dblp.org}
}

@article{hayes2022practical,
  author       = {Conor F. Hayes and
                  Roxana Radulescu and
                  Eugenio Bargiacchi and
                  Johan K{\"{a}}llstr{\"{o}}m and
                  Matthew Macfarlane and
                  Mathieu Reymond and
                  Timothy Verstraeten and
                  Luisa M. Zintgraf and
                  Richard Dazeley and
                  Fredrik Heintz and
                  Enda Howley and
                  Athirai A. Irissappane and
                  Patrick Mannion and
                  Ann Now{\'{e}} and
                  Gabriel de Oliveira Ramos and
                  Marcello Restelli and
                  Peter Vamplew and
                  Diederik M. Roijers},
  title        = {A practical guide to multi-objective reinforcement learning and planning},
  journal      = {Auton. Agents Multi Agent Syst.},
  volume       = {36},
  number       = {1},
  pages        = {26},
  year         = {2022},
  url          = {https://doi.org/10.1007/s10458-022-09552-y},
  doi          = {10.1007/S10458-022-09552-Y},
  bibsource    = {dblp computer science bibliography, https://dblp.org}
}

@inproceedings{wu2023finegrained,
  author       = {Zeqiu Wu and
                  Yushi Hu and
                  Weijia Shi and
                  Nouha Dziri and
                  Alane Suhr and
                  Prithviraj Ammanabrolu and
                  Noah A. Smith and
                  Mari Ostendorf and
                  Hannaneh Hajishirzi},
  editor       = {Alice Oh and
                  Tristan Naumann and
                  Amir Globerson and
                  Kate Saenko and
                  Moritz Hardt and
                  Sergey Levine},
  title        = {Fine-Grained Human Feedback Gives Better Rewards for Language Model
                  Training},
  booktitle    = {Advances in Neural Information Processing Systems 36: Annual Conference
                  on Neural Information Processing Systems 2023, NeurIPS 2023, New Orleans,
                  LA, USA, December 10 - 16, 2023},
  year         = {2023},
  url          = {http://papers.nips.cc/paper\_files/paper/2023/hash/b8c90b65739ae8417e61eadb521f63d5-Abstract-Conference.html},
  bibsource    = {dblp computer science bibliography, https://dblp.org}
}

@inproceedings{wang2024armorm,
  author       = {Haoxiang Wang and
                  Wei Xiong and
                  Tengyang Xie and
                  Han Zhao and
                  Tong Zhang},
  editor       = {Yaser Al{-}Onaizan and
                  Mohit Bansal and
                  Yun{-}Nung Chen},
  title        = {Interpretable Preferences via Multi-Objective Reward Modeling and
                  Mixture-of-Experts},
  booktitle    = {Findings of the Association for Computational Linguistics: {EMNLP}
                  2024, Miami, Florida, USA, November 12-16, 2024},
  series       = {Findings of {ACL}},
  volume       = {{EMNLP} 2024},
  pages        = {10582--10592},
  publisher    = {Association for Computational Linguistics},
  year         = {2024},
  url          = {https://doi.org/10.18653/v1/2024.findings-emnlp.620},
  doi          = {10.18653/V1/2024.FINDINGS-EMNLP.620},
  bibsource    = {dblp computer science bibliography, https://dblp.org}
}

@inproceedings{moskovitz2024confronting,
  author       = {Ted Moskovitz and
                  Aaditya K. Singh and
                  DJ Strouse and
                  Tuomas Sandholm and
                  Ruslan Salakhutdinov and
                  Anca D. Dragan and
                  Stephen Marcus McAleer},
  title        = {Confronting Reward Model Overoptimization with Constrained {RLHF}},
  booktitle    = {The Twelfth International Conference on Learning Representations,
                  {ICLR} 2024, Vienna, Austria, May 7-11, 2024},
  publisher    = {OpenReview.net},
  year         = {2024},
  url          = {https://openreview.net/forum?id=gkfUvn0fLU},
  bibsource    = {dblp computer science bibliography, https://dblp.org}
}

@article{huang2026focal,
  author       = {Yu Huang and
                  Zihua Zhao and
                  Zhaoxin Huan and
                  Wanli Gu and
                  Feng Hong and
                  Xinmu Ge and
                  Lin Yuan and
                  Weichang Wu and
                  Qiang Hu and
                  Xiaolu Zhang and
                  Jun Zhou and
                  Jiangchao Yao},
  title        = {Focal Reward: Balanced Reinforcement Learning under Rubric-Based Rewards},
  journal      = {CoRR},
  volume       = {abs/2605.26579},
  year         = {2026},
  url          = {https://doi.org/10.48550/arXiv.2605.26579},
  doi          = {10.48550/ARXIV.2605.26579},
  eprinttype   = {arXiv},
  eprint       = {2605.26579},
  bibsource    = {dblp computer science bibliography, https://dblp.org}
}

@misc{qwen3.6-27b,
    title  = {{Qwen3.6-27B}: Flagship-Level Coding in a {27B} Dense Model},
    author = {{Qwen Team}},
    month  = {April},
    year   = {2026},
    url    = {https://qwen.ai/blog?id=qwen3.6-27b}
}

@article{mistral,
  author  = {Albert Q. Jiang and Alexandre Sablayrolles and Arthur Mensch and
             Chris Bamford and Devendra Singh Chaplot and Diego de las Casas and
             Florian Bressand and Gianna Lengyel and Guillaume Lample and
             Lucile Saulnier and L{\'e}lio Renard Lavaud and Marie-Anne Lachaux and
             Pierre Stock and Teven Le Scao and Thibaut Lavril and Thomas Wang and
             Timoth{\'e}e Lacroix and William El Sayed},
  title   = {Mistral 7{B}},
  journal = {CoRR},
  volume  = {abs/2310.06825},
  year    = {2023},
  url     = {https://arxiv.org/abs/2310.06825},
}

@article{vllm,
  author  = {Woosuk Kwon and Zhuohan Li and Siyuan Zhuang and Ying Sheng and
             Lianmin Zheng and Cody Hao Yu and Joseph E. Gonzalez and Hao Zhang and
             Ion Stoica},
  title   = {Efficient Memory Management for Large Language Model Serving with {PagedAttention}},
  journal = {CoRR},
  volume  = {abs/2309.06180},
  year    = {2023},
  url     = {https://arxiv.org/abs/2309.06180},
}
